\documentclass[%
reprint,
superscriptaddress,
floatfix,
amsmath,
amssymb,
aps,
notitlepage
]{revtex4-1}
\usepackage[hidelinks=true]{hyperref}
\usepackage{xcolor}
\definecolor{greenish}{rgb}{0.13,0.58,0.16}
\definecolor{reddish}{RGB}{174,12,48}
\definecolor{blueish}{rgb}{0.12, 0.56, 1.0}
\definecolor{magenta}{rgb}{0.8, 0.0, 0.8}

\hypersetup{
colorlinks   = true, 
urlcolor     = greenish, 
linkcolor    = magenta, 
citecolor   = blueish 
}
\usepackage{setspace}
\usepackage{siunitx}
\usepackage{subcaption}
\usepackage{graphicx}
\usepackage{dcolumn}
\usepackage{bm}

\def\grad{\nabla}

\def\nh{\hat{\mathbf{p}}}
\def\r{\mathbf{r}}

\def\cm{\text{cm}}

\def\u{\mathbf{u}}
\def\e{\hat{\mathbf{e}}}

\def\O{\mathcal{O}}

\def\C{\mathsf{C}}

\def\x{\mathbf{x}}

\def\s{{\rm s}}

\def\kh{\hat{k}}
\def\T{\mathsf{E}}

\def\vh{\mathsf{V}}
\def\dh{\mathsf{D}}

\def\ss{\text{ss}}

\def\grad{\nabla}

\def\nh{\hat{\mathbf{p}}}
\def\r{\mathbf{r}}

\def\cm{\text{mm}}

\def\u{\mathbf{u}}
\def\e{\hat{\mathbf{e}}}

\def\O{\mathcal{O}}

\def\C{\mathsf{C}}

\def\x{\mathbf{x}}

\def\s{\text{s}}

\def\kh{\hat{k}}

\def\nh{\hat{\mathbf{n}}}
\def\d{\text{d}}
\def\ph{\hat{\mathbf{p}}}
\def\nh{\hat{\mathbf{n}}}
\def\rh{\hat{\mathbf{r}}}
\def\e{\text{e}}

\def\x{\mathbf{x}}
\def\R{\mathcal{R}}

\def\gt{\mathsf{G}}

\def\pst{\tilde{\psi}}
\def\rt{\tilde{r}}
\def\Z{\mathbb{Z}}
\def\ss{\text{ss}}

\def\s{\text{s}}
\def\T{\mathsf{K}}
\def\o{\text{o}}
\def\L{\mathsf{L}}
\def\ndr{\mathsf{r}}
\def\K{\mathsf{K}}
\def\R{\mathsf{R}}

\definecolor{apricot}{rgb}{0.98, 0.81, 0.69}
\definecolor{mulberry}{rgb}{0.77, 0.29, 0.55}
\definecolor{ruby}{rgb}{0.88, 0.07, 0.37}
\definecolor{forestgreen}{rgb}{0.13, 0.55, 0.13}

\newcommand{\robDes}{S6 }
\newcommand{\SISRAnt}{S2 }
\newcommand{\dynamic}{S2B }
\newcommand{\contmodel}{S4 }
\newcommand{\SIFigOne}{S1 }
\newcommand{\SIFigTrap}{S2 }
\newcommand{\clusterNumber}{S6}
\newcommand{\simDet}{S5 }

\newcommand{\rhos}{6 }
\newcommand{\TwoDsimul}{4}
\newcommand{\rantResults}{3 }

\begin{document}
	\title{Collective phototactic robotectonics}
	\author{Fabio Giardina}
	\thanks{Equal contribution}
	\affiliation{School of Engineering and Applied Sciences, Harvard University, Cambridge, MA 02138.}
	\author{S Ganga Prasath}
	\thanks{Equal contribution}	
	\affiliation{School of Engineering and Applied Sciences, Harvard University, Cambridge, MA 02138.}
	\author{L Mahadevan}
	\email{lmahadev@g.harvard.edu}
	\affiliation{School of Engineering and Applied Sciences, Harvard University, Cambridge, MA 02138.}
	\affiliation{Center for Brain Science, Harvard University, Cambridge, MA 02138.}
	\affiliation{Department of Physics, Harvard University, Cambridge, MA 02138.}
	\affiliation{Department of Organismic and Evolutionary Biology, Harvard University, Cambridge, MA 02138.}


	\begin{abstract}
	Cooperative task execution, a hallmark of eusociality, is enabled by local interactions between the agents and the environment through a dynamically evolving communication signal. Inspired by the collective behavior of social insects whose dynamics is modulated by interactions with the environment, we show that a robot collective can successfully nucleate a construction site via a trapping instability and cooperatively build organized structures. The same robot collective can also perform \textit{de}-construction with a simple change in the behavioral parameter. These behaviors belong to a two-dimensional phase space of cooperative behaviors defined by agent-agent interaction (cooperation) along one axis and the agent-environment interaction (collection and deposition) on the other. Our behavior-based approach to robot design combined with a principled derivation of local rules enables the collective to solve tasks with robustness to a dynamically changing environment and a wealth of complex behaviors.
	\end{abstract}

	\pacs{Valid PACS appear here}
	\maketitle
	
\begin{figure*}[t!]
\centering
\includegraphics[width=\textwidth]{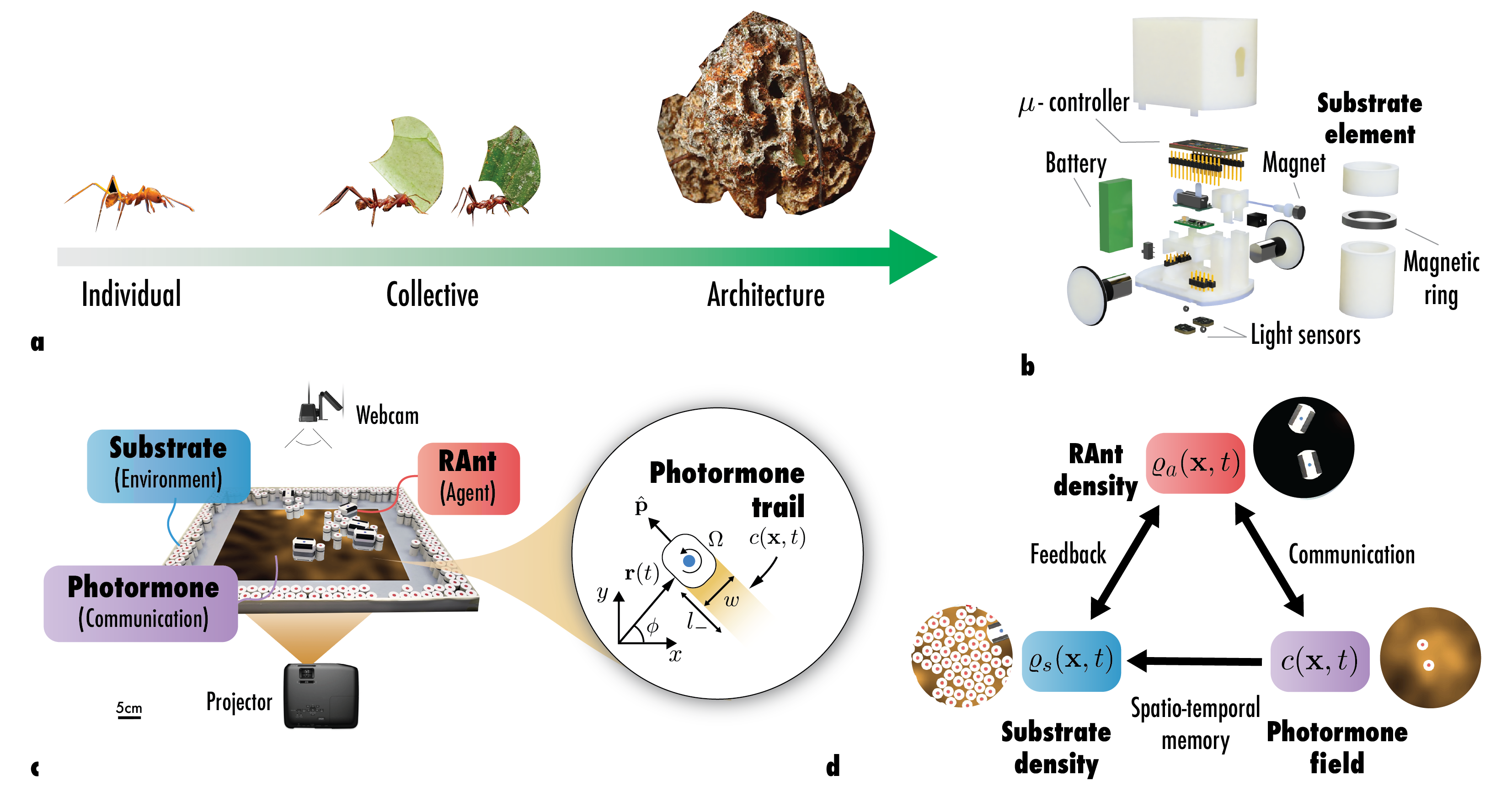}
\caption{\textbf{\textsf{Construction of structures in social insects and robots}}\ $(a)$\ Social insects cooperatively construct architectures that are much larger than their size because of their collective effort. Here we see leafcutter ants collectively foraging leaves into their nest. $(b)$ RAnts are autonomous wheeled robots and are each equipped with two light sensors to measure the local photormone intensity (gradient information) and a servomotor equipped with a magnet to transport substrate elements. The substrate element has a magnetic ring that enables easy attachment and detachment from the RAnts. $(c)$ Robotic platform for construction and de-construction experiments where we have the RAnts that communicate with each other via \textit{photormone field} that they generate along their trajectory. A webcam tracks the position of the RAnts and is used to compute the photormone production via Eq.~\ref{eq:photoDless} which is then displayed on the working surface of the robots using a projector. Each RAnt is described by its position $\r(t)$, orientation $\ph$ (see Eqs.~\ref{eq:pos}-\ref{eq:orn}) and it leaves behind a trail of photormone $c(\x, t)$ of width $w$ (described in Eq.~\ref{eq:photoDless}). It also responds to gradients in photormone along its path by rotating at a speed $\Omega$ proportional to the gradient along the normal direction. $l_-$ is the length-scale over which the photormone decays and $l_\s$ is the distance between the light sensors (also in Eq.~\ref{eq:photoDless}). $(d)$ Triadic interaction between three relevant variables in the model and experiments: RAnt density $\varrho_a (\x, t)$ representing the agents, photormone field $c (\x, t)$ used for communication and the substrate density $\varrho_s (\x, t)$ (ref. Eqs.~\ref{eq:rhor}-\ref{eq:rhos}) that acts as the environment. Agents interact with each other through a photormone field and interact with the substrate by carrying and placing them. The location of high photormone concentration acts as spatio-temporal memory where construction happens and gets reinforced by repeated RAnt visits.}
\label{fig:intro}
\end{figure*}

The solution of complex problems on scales much larger than the size of an individual, in both natural~\cite{gordon1999ants, holldobler1990ants, peleg2018collective, ocko2014collective, peters2019collective, heyde2021self, feinerman2018physics, camazine2020self} and artificial systems~\cite{giomi2013swarming,wang2022robots,pfeifer2007self, petersen2019review}, often requires the cooperative effort of a collective. An example is the collective construction task in leafcutter ants where they cut leaves and carry them from the plant to the nest in a manner that reduces the individual effort while the collective effort ensures their survival and existence (see Fig.~\ref{fig:intro}$(a)$). In addition to understanding how organisms communicate, cooperate and carry out functional tasks, it is useful to ask how easy or difficult it is to synthesize these tasks in-silico or in biomimetic systems~\cite{rahwan2019machine}. One difficulty is that the participating agents in a collective must interact with and modify the environment around them while simultaneously being influenced by it~\cite{perna2017social, camazine2020self, Prasath2021cooperative}. Any collective task that alters the environment in a structured manner thus requires the agent to $(i)$ communicate with other agents and recruit them from the colony into the quorum, and $(ii)$ physically interact and move material based on the state of the environment.

Several natural systems~\cite{theraulaz1999brief,perna2017social, heyde2021self,holland1999stigmergy,camazine2020self} use \textit{stigmergy} as a recruitment strategy, wherein the agents leave signals such as pheromones in the environment. This serves as a spatio-temporal memory to harness more individuals into the collective, and has inspired the design of synthetic systems~\cite{fujisawa2008dependency, mayet2010antbots, khaliq2014stigmergic, valentini2018kilogrid, ravankar2016bio, llenas2018quality, reina2015augmented, sugawara2004foraging, garnier2007alice, arvin2015cosvarphi, na2019extended, wang2021emergent}. Then, task execution using stigmergy can be thought of as a triadic interaction between three relevant variables: the agents, the stigmergic communication field, and the environment (see Fig.~\ref{fig:intro}$(d)$) which vary spatio-temporally towards task execution. An important aspect here is to ask how individuals can reach a consensus that is robust to changes in both their behaviors and environmental changes, and yet be collectively flexible enough to not be limited to a single task, e.g. be capable of construction and its inverse, \textit{de}-construction. This naturally raises two questions: $(i)$ how do we design a set of microscopic behavioral rules at the level of an individual agent that leads to the emergence of robust and flexible task completion? $(ii)$ how might we engineer a synthetic system to explore the transition from individual to collective and flexible task execution?

Here we investigate these questions by using a robotic collective similar to that used in a number of recent studies~\cite{rubenstein2012kilobot,rubenstein2014programmable,aguilar2018collective,wang2022robots} and our recent study \cite{Prasath2021cooperative} where we investigated a specific excavation behavior exhibited by black carpenter ants. In contrast to our previous work, here we derive a principled approach for the synthesis of a broader class of cooperative behaviors: collective architecture.
We start by incrementally designing and implementing simple behavioral rules for how robots sense and move in response to each other and their local environment, represented here by a combination of haptotactic and phototactic stimuli, to mimic what is seen in ants~\cite{gordon1999ants, perna2017social, camazine2020self}. The robots in the collective communicate with each other via a light field, similar to a pheromone field, that they generate along their trajectory. This light signal affects their behavior, and results in the robots manipulating the environment. We show that robots can reach a consensus on a task such as construction by collectively creating a nucleation site for construction via a trapping instability where the robots trap themselves and others in the communication field (similar to pheromones in social insects). By varying a single microscopic behavioral parameter that modifies the way in which robots interact with their environment, we see that the robots can switch from construction to de-construction, and thus navigate the complex phase space of cooperative behaviors via simple changes in agent behavior. This leads to cooperative task completion that is robust even in uncertain environments that are molded and in turn mold the collective behavior.
\begin{figure}[t!]
\centering
\includegraphics[width=0.49\textwidth]{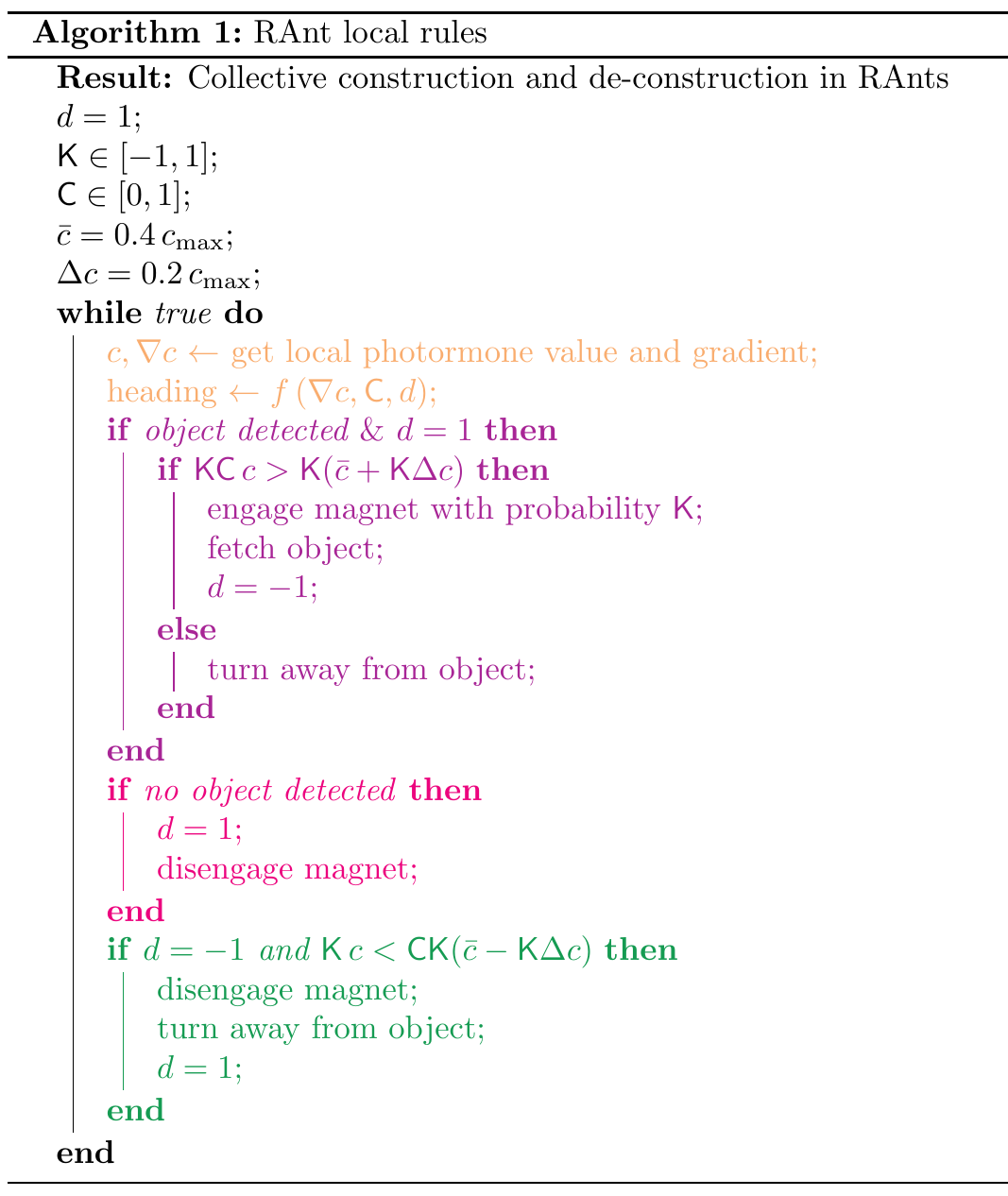}
\caption{\textbf{\textsf{Pseudocode of internal robot rules}} The codes starts by initializing all relevant parameters, namely the movement direction $d$ (1 for forward, -1 for backward), the deposition rate $\mathsf{K}$, the cooperation parameter $\mathsf{C}$, and the threshold parameters $\bar{c}$ and $\Delta c$. The main loop can be broken down into four parts. The {\color{apricot} \bf{first}} part of the loop measures the local photormone concentration $c$ and its gradient $\nabla c$ and updates the robot's heading as a function of $\nabla c$,  $\mathsf{C}$ and $d$. The {\color{mulberry} \bf{second}} part handles the obstacle detection and fetches detected obstacles if the photormone threshold is reached, and avoids obstacles otherwise. The {\color{ruby} \bf{third}} part ensures detachment from objects if nothing is detected. Finally, the {\color{forestgreen} \bf{fourth}} part of the loop handles release of captured objects when the measured photormone falls below the lower threshold.}\label{fig:code}
\end{figure}

\noindent \textit{Robotic ants and their assembly:} To mimic construction tasks such as those performed by leafcutter ants who cut and move leaves to their nests while being guided by pheromones (see Fig.~\ref{fig:intro}$(a)$), we construct a robotic system based on our recent study~\cite{Prasath2021cooperative}. In order to physically realize the capability to sense a scalar external field and its gradients, we use phototactic sensors, and to enable the robots to change their environment, we provide the robots the ability to physically lift and move cylindrical objects. These capabilities allow the robots to be able to (ref. SI sec.~\clusterNumber): $(i)$ produce a communication signal/field that they can leave in the environment and that can be sensed by other agents; $(ii)$ detect and move the substrate material used for construction; $(iii)$ change their behavioral rules upon sensing the communication signal. Our robotic platform for collective behavior that can achieve this is shown in Fig.~\ref{fig:intro}$(c)$, and consists of an arena in which our wheeled \textit{RAnts} (Robot Ants) can move, while producing and sensing a \textit{photormone} field, realized by a light projection onto a translucent screen. The RAnts are battery-driven and have an on-board microcontroller encoded with behavioral rules to sense, move and build/unbuild (see Fig.~\ref{fig:code}) (see SI sec.~\robDes for more details). Sensing is enabled by two light sensors that can detect the intensity of the photormone field and its gradient perpendicular to their direction of travel (see Fig.~\ref{fig:intro}$(b)$ and SI sec.~\robDes for more details). The RAnts can move and turn using two independently-controlled wheels, and can detect obstacles in front of them with an infrared sensor. They are also equipped with an extensible magnet that can attach and detach from cylindrical segments, that serve as architectural building blocks. The blocks are made out of PVC pipes, each with a magnetic ring inclusion. The combination of internal behavioral rules (example shown in Fig.~\ref{fig:code}) and the embodied collective interactions between the RAnts and the environment leads to the emergence of photormone patterns, coordinated movement and collective building/unbuilding. This requires the implementation of a triadic interaction of agents, communication field, and environment by starting at the level of individual behavioral rules. \\

\begin{figure*}[t!]
\centering
\includegraphics[width=\textwidth]{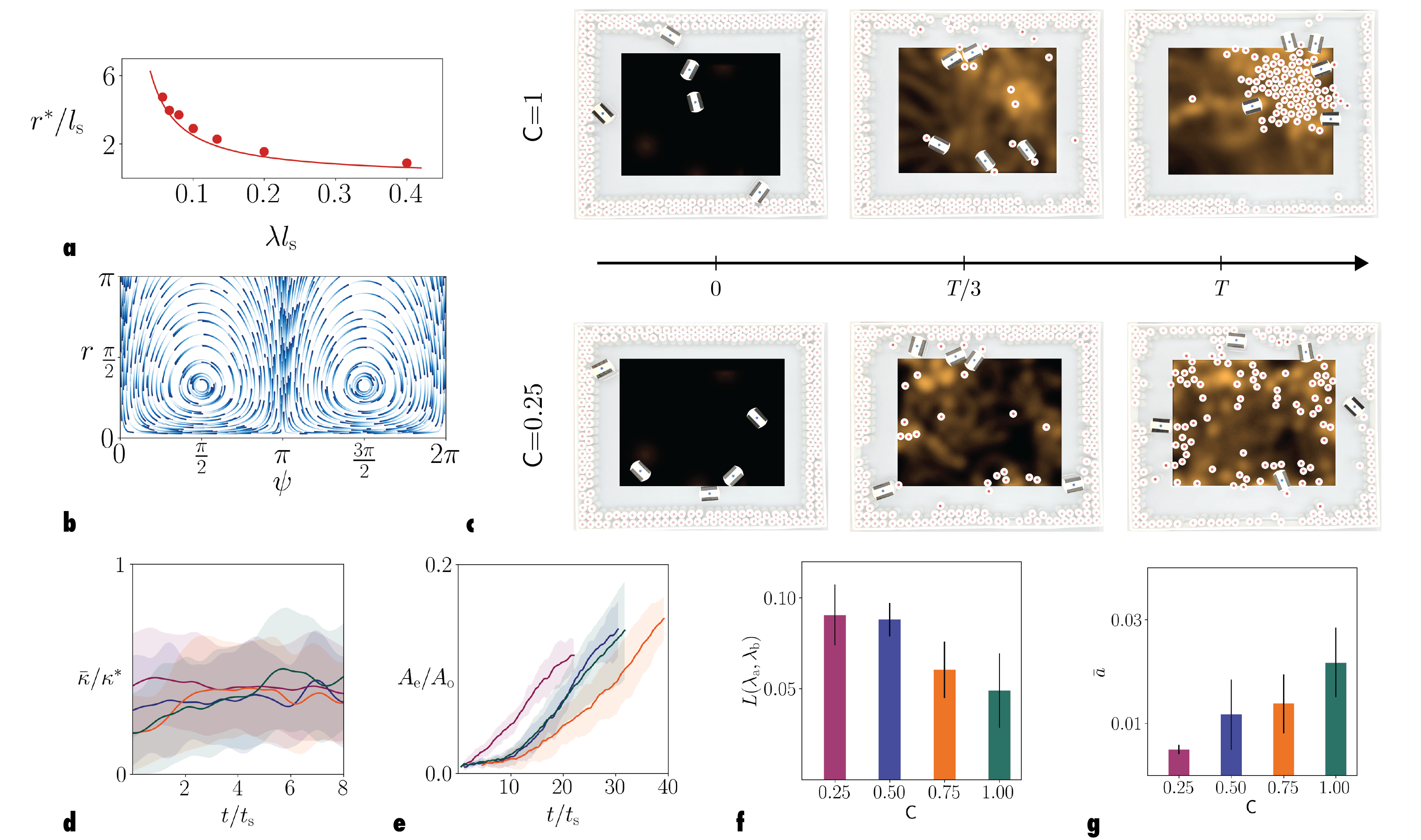}
\caption{\textbf{\textsf{Collective robotic construction}} \ $(a)$ Fixed point measured in experiments by projecting a field of constant gradient where $r^* = 1/\lambda$ discussed in Eqs.~\ref{eq:pos}-\ref{eq:orn}. $(b)$ Phase-space $(\psi, r)$ representing the RAnt dynamics with fixed points $(\psi^*, r^*)$ (also from Eqs.~\ref{eq:pos}-\ref{eq:orn}). $(c)$ Series of snapshots of robot construction progress at different times for cooperation parameters $\C=0.25$ and $\C=1$. $(d)$ Non-dimensional average curvature of the RAnts' trajectories, $\bar{\kappa}/\kappa^*$, with $\kappa^*$ being the inverse of the trapping radius, as a function of normalized time and for different cooperation parameters $\C$. $t_\text{s}$ is the time it takes for a RAnt to cross the arena. $(e)$ Normalized covered area of deposited substrate $A_\text{e}/A_\o$ ($A_\o$ construction arena area) as a function of time and for different cooperation parameters $\C$. $(f)$ Average cluster area relative to construction area $\bar{a}=(n_{\textrm{c}} A_\o)^{-1} \sum_{i=1}^{n_\textrm{c}} A_i$ as a function of the cooperation parameter $\C$. $n_{\textrm{c}}$ is the number of clusters, $A_0$ the area of the construction arena, and $A_i$ the area covered by the $i$th cluster. $(g)$ Circumference $L$ of the ellipse defined by the eigenvalues $\lambda_{\rm{a}}, \lambda_{\rm{b}}$ of the sample covariance matrix of the substrate elements in the construction area as a function of the cooperation parameter $\C$.}
\label{fig:rantResults}
\end{figure*}

\noindent \textit{Single RAnt in static photormone field:} The amplitude of response of a robot to the external photormone field can be quantified in terms of a non-dimensional number we call the cooperation parameter, $\C \in [0,1]$; a random walk corresponds to $\C=0$, while perfect gradient-following corresponds to $\C=1$. When $\C=1$, each RAnt's position $\r(t)$ and orientation $\ph(t)$ (ref. Fig.~\ref{fig:intro}$(c)$ inset) evolves according to the following equations: 
\begin{align}
\dot{\r} =& \ v_o \ph, \label{eq:rSing}\\
\dot{\ph} =& \ \Omega \times \ph. \label{eq:pSing}
\end{align}
Here the orientation is given by $\ph = (\cos \theta, \sin \theta)$, $\theta$ is the heading angle, $v_o$ is the base speed of the RAnt. RAnts can sense the gradient in a photormone field $c(\x,t)$ and that causes the RAnts to reorient at a rate $\Omega = G(\grad c \cdot \nh) \hat{z}$ where $\nh = (-\sin \theta, \cos \theta)$, $\hat{z}$ is the out-of-plane direction and $G$ the rotational gain. In a constant gradient with $\grad c = -\lambda \rh$, writing $\rh = (\cos \phi, \sin \phi)$ with $\phi$ being the polar angle relative to a laboratory axis, the rotation rate becomes $\Omega = G \lambda \sin(\theta-\phi) \hat{z}$. The dynamical equations of motion can be written in non-dimensional form by scaling lengths using the intrinsic length-scale from the prescribed gradient $\lambda^{-1}$ and time using the intrinsic time-scale $1/(v_o \lambda)$. In the transformed variables $r \rightarrow \tilde{r}/\lambda, t \rightarrow \tilde{t}/(v_o \lambda)$, the evolution Eqs.~\ref{eq:rSing},~\ref{eq:pSing} become (after dropping tildes):
\begin{align}
\dot{r} =& \ \cos \psi, \label{eq:pos}\\
\dot{\psi} =& \ \bigg( \gt - \frac{1}{r} \bigg) \sin \psi, \label{eq:orn}
\end{align}
where $\psi = (\theta - \phi)$, and $\gt = G/v_o$ is the photormone sensitivity. We see that Eqns.~\ref{eq:pos}, \ref{eq:orn} have a fixed point in the dynamics at $r^* = 1/\gt, \psi^* = \pi/2$ and corresponds to a circular trajectory with constant radius $r=r^*$ as seen in Fig.~\ref{fig:rantResults}$(b)$ in the $(\psi, r)$-plane. 

To test this simple prediction we use our robotic platform (shown in Fig.~\ref{fig:intro}$(c)$) to address how RAnts move in a  stationary photormone field with a constant gradient, $c = -\lambda r$. Indeed a RAnt moves in a circle of radius $1/\gt$ (see SI Video 1). In Fig.~\ref{fig:rantResults}$(a)$ we show that the experimentally measured $r^*$ vs $\lambda$, scaled using the distance between light sensors $l_\s$, is consistent with the scaling result $r^* \sim \gt^{-1}$. Perturbations of the RAnt dynamics with initial conditions away from this fixed point result in quasi-periodic dynamics with precession of the RAnt orientation as shown in the phase-space of the RAnts defined in the $(\psi, r)$-plane (see Fig.~\ref{fig:rantResults}$(b)$ and ref. SI sec.~\SISRAnt for further details).

\noindent \textit{RAnts in a dynamic photormone field:} To allow the photormone field to evolve in response to the motion of RAnts, we now assume that the agents can actively produce photormone at a rate $k_+$, and that the field decays over time at a constant rate $k_-$. The photormone field $c(\x,t)$ then evolves according to
\begin{align}
\partial_t c(\x,t) =& \ k_+ \varrho_a[\r, w] - k_- c(\x,t), \label{eq:photoDless}
\end{align}
where $\varrho_a = \{ 1$ if $|\x-\r|^2 - w^2 \leq 0$; $0$ if $|\x-\r|^2 - w^2 > 0 \}$ is the RAnt density that has a unit magnitude in a circle surrounding the agent with a width $w$ where the photormone is produced, zero elsewhere. Equations~\ref{eq:rSing},~\ref{eq:pSing},~\ref{eq:photoDless} complete the formulation of a RAnt that deposits a decaying photormone field and in turn responds to it.

The dimensional Eq.~\ref{eq:photoDless} has the 3 length-scales: $l_{\s}, w$ and $l_- (= v_o/k_-)$ where $l_{\s}$ is the physical distance between the light sensors on the RAnts, $w$ is radius over which photormone is produced and $l_-$ is the length scale over which the RAnts travel before the photormone they generate decays e-fold (see Fig.~\ref{fig:intro}$(c)$ for a schematic). We can define 2 non-dimensional numbers using these length-scales as: $\L_- = l_-/l_\s, \L_w = w/l_\s$. In the two limits of scale-separation given by $\L_- \ll \L_w \sim \O(1)$ and $\L_w \sim \O(1) \ll \L_-$ we shall see that for cooperation parameter $\C=1$, the RAnt is trapped in its photormone field, whereby its trajectories are bound orbits (which we call \textit{traps}).

When the length-scale over which pheromone decays $l_-$ is small relative to the width of photormone generation $w$, this corresponds to $\L_- \ll \L_w$. The gradient measured by the sensor is then determined by the size over which they generate photormone i.e. $w$. In this limit the maximum gradient measured by a RAnt as it generates photormone is at the specific location $\ndr^* = (\L_w + 1)/4$ where $\ndr=r/l_\s$ (see SI sec.~\SISRAnt for details) and is expected to be the radius of the bound orbit. We see that the radius of self-trapping is purely geometric when we are in the limit of $\L_- \ll \L_w$ as the trapping radius depends only on the size of photormone production $w$ and distance between sensors. However, for this distance to be the radius of trapping realizable in experiment we need to ensure that the RAnts are capable of making such a turning radius. From our earlier analysis for constant gradient in Eqs.~\ref{eq:pos},~\ref{eq:orn} we know that the critical radius is $\ndr_c = 1/(\gt\grad c)$. Therefore, in order for the trapping radius $\ndr^*$ to be realizable, we require that it be greater or equal to the critical turning radius $\ndr_c$ i.e. $\ndr^* \geq \ndr_c$. We can identify the boundary of self-trapping using the conditions $\ndr^*=\ndr_c$. This provides us with the critical gain, $\gt_c$ required for the RAnts to self-trap at a radius $\ndr^*$ for a given $\L_w$. 

We arrive at the expression for $\gt_c$ by first finding the steady state photormone intensity assuming a RAnt travels along a circle of radius $\ndr^*$ by solving Eq.~\ref{eq:photoDless}. Using the steady state profile, we calculate the gradient around this trajectory subject to the compatibility turning condition, $\ndr^* \geq \ndr_c$. We briefly outline the calculation here and the details are in SI sec.~\dynamic. For a given $\ndr^*$, the steady-state photormone concentration of Eq.~\ref{eq:photoDless} closer to the center of the trap is expected to be a constant $c=k_+/k_-=\kh_\pm$. This is because the limit $\L_w \gtrsim 1$ implies that the photormone production diameter $w$ is large enough to cover the circle drawn by the trajectory of the sensor. Hence, for the sensor closer to the center, the measured photormone value is governed by Eq.~\ref{eq:photoDless} with $\varrho_a = 1$: $\dot{c} = \kh_\pm - c$, where we have non-dimensionalized Eq.~\ref{eq:photoDless} by using $(1/k_-)$ as the time-scale and $l_s$ as the length-scale. The outer sensor on the other hand travels along a circle of radius $\R=(\ndr^*+1)/2$ (see SI Fig.~\SIFigOne $(d)$ for a schematic) and ``sees'' photormone production only for a fraction of the inner sensor. After solving for the photormone field in Eq.~\ref{eq:photoDless}, we get $\grad c = [ c(\R) - \kh_\pm]$. The value of photormone at the outer sensor converges to
\[
c(\R) = \kh_\pm \bigg[ 1-\exp \bigg( \frac{\tau_1}{2} \bigg) \frac{\exp(\tau_2) - 1}{\exp(\tau_1 + \tau_2) - 1} \bigg],
\]
where $\tau_1$ is the time over which RAnts produce photormone as they traverse the circle and $\tau_1 + \tau_2 = 2\pi \ndr^*/\L_-$ is the time taken to go around the circle $\ndr^*$ once. Ultimately we arrive at the expression for critical gain of each RAnt to trap given by $\gt_c = 1/[\kh_\pm - c(\R)]$. On the other hand when $\L_w \ll \L_-$, the geometry associated with the generation of photormone no longer affects the trapping dynamics. In this limit we expect the radius of trapping to be large and from a similar analysis we find that $\ndr^* \approx \L_w \gt \kh_\pm/2\pi$ (see SI sec.~\dynamic for details). The trapping conditions for a given field dynamics i.e. fixed $\kh_\pm$ thus depends on the photormone sensitivity and agent's mobility through $\gt$, the geometry of the sensor arrangement and photormone production through $\L_w, \L_-$.

To summarize, when $\L_- \ll \L_w$ the trapping radius is $\ndr^* = (\L_w + 1)/4$ as long as $\ndr^* > \ndr_c$. We can ensure this by choosing $\gt_c = 1/[\kh_\pm - c(\R)]$. On the other hand when $\L_w \ll \L_-$, the trapping radius is $\ndr^* \approx \L_w \gt \kh_\pm/2\pi$. We have described here the trapping conditions required for a single RAnt while the condition for multiple RAnts in a collective field is discussed in the SI sec.~\dynamic (see also  Fig~\SIFigTrap $(a)$). Trapping stops a RAnt from exploring the space it operates in and causes a build-up of photormone in a confined region of space. We can avoid trapping of a single agent by setting the internal gain $\gt_c$ below the theoretical predictions and thus have control over the number of agents that must be within the trapping radius to start a trap (which we tested in experiments, see SI Video 2). Our results are qualitatively similar to earlier studies on trapping in single agent~\cite{tsori2004self,taktikos2011modeling,kranz2016effective} as well multiple agents~\cite{hillen2009user} that show similar trapping mechanisms.  As we shall see later, the trapping instability we have analyzed here acts as a mechanism for spontaneous seed formation for construction: a trap made by a number of agents forms an effective attractor that can be exploited to spatially coordinate the construction effort.\\

\begin{figure*}[t!]
\centering
\includegraphics[width=0.9\textwidth]{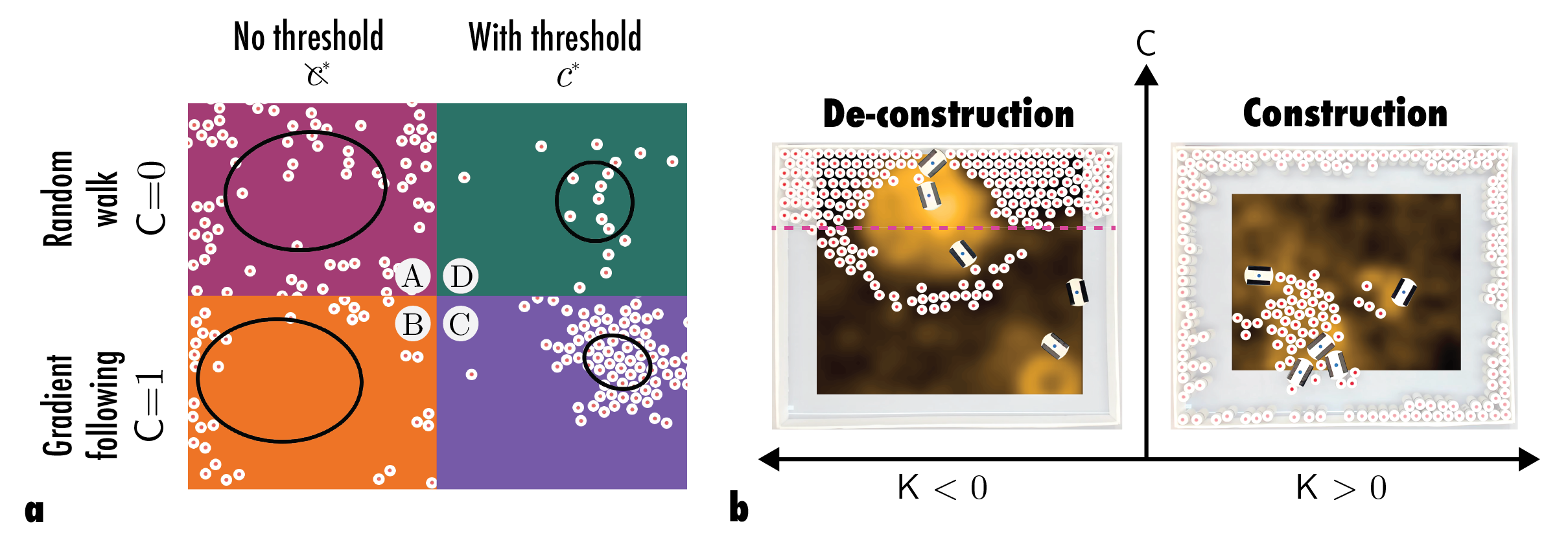}
\caption{\textbf{\textsf{Robustness and unified phase space}} $(a)$ Distribution of substrate elements at the end of experiments for 4 scenarios: (A) with no threshold and phototaxis, $c^* = 0, \C=0$;  (B) with no threshold and strong phototaxis,  $c^* = 0, \C=1$; (C) with threshold and strong phototaxis, $c^* \neq 0, \C=1$; (D) with threshold and no phototaxis, $c^* \neq 0, \C=0$ (see SI Fig.~\clusterNumber$(b)$ for a quantitative comparison). The black ellipses represent the sample covariance of the substrate elements in the construction area. We see that the clustering is densely packed only when both threshold and phototaxis is present, $c^* \neq 0$ and $\C=1$. $(b)$ Distribution of substrate elements at the end of experiments under high cooperation, $\C = 1$ for construction $\K > 0$ and de-construction $\K < 0$. The tasks exhibited by the RAnt collective is captured in this unified phase space represented by $\C$ vs $\K$.}
\label{fig:robust}
\end{figure*}

\noindent \textit{Algorithm for collective robotectonics:} Two important mechanisms intrinsic to RAnts facilitate collective construction: $(i)$ ability to follow gradients or phototaxis, $(ii)$ a thresholded response to photormone which sets the criteria for consensus of the agents. While the phototaxis ability is important for a directed motion towards the construction seed, the threshold criteria for consensus helps to coordinate the collective effort by using the photormone concentration as a spatio-temporal memory of the previous activity of the collective. How can we encode these behaviors into a set of rules for the RAnts to follow?

In Fig.~\ref{fig:code}, we provide a behavioral algorithm for collective construction. A short summary of the algorithm follows: $(i)$ the main loop, \textit{set heading}, determines the rotation of the robot which is a function of the local photormone gradient, $\nabla c$, the cooperation parameter $\C$ and the direction of travel $d$; $(ii)$ simultaneously the robot uses two thresholds (assuming for simplicity a deposition rate of $\mathsf{K}$=1) $\bar{c}\pm \Delta c$ for collection and deposition of a detected object. These thresholded rules ensure that robots coordinate their collection of substrate at high values of self-generated photormone field (where robots \textit{have} been recently), while their deposition occurs at low values (where robots \textit{have not} been recently). To allow for nucleation of a construction site via a trapping instability, we chose a rotational gain $\gt$ that permits the formation of traps in the presence of 5 or more robots. We found this value to facilitate stable yet adaptable nucleation sites that evolve dynamically with the state of the robots and the architectural site.
 
In Fig.~\ref{fig:rantResults}$(c)$ we show how the magnitude of the cooperation parameter, $\C \in [0.25, 1]$ (implemented through phototaxis strength) leads to varying architectural patterns while keeping the deposition rate constant at $\K = 1$. Initially, RAnts attach to the first substrate element they encounter and then perform a mixture of a random walk (ref. SI Video 3) and gradient ascent according to their cooperation parameter (see Fig.~\ref{fig:rantResults}$(c)$ left panels). Once a RAnt with an attached substrate element detects a photormone concentration above the detachment threshold, the element is dropped. We see that higher cooperation, $\C \geq 0.5$ leads to higher average curvature, $\bar{\kappa}$ (see Fig.~\ref{fig:rantResults}$(d$)) leading to more and more of the RAnts being at the same location. This also results in increase in the rate of construction as shown in Fig.~\ref{fig:rantResults}$(e)$. When $\C$ is large we see the formation of a single large isotropic cluster, corresponding to the location with a clear peak in the photormone concentration  
(ref. Fig.~\ref{fig:rantResults}$(f)$, SI sec. S7 for details of quantification of the shape of the clusters). As $\C$ is reduced we see a more dispersed deposition pattern (ref. Fig.~\ref{fig:rantResults}$(c)$) leading to a larger average area covered by the formed clusters, $\bar{a}$ (shown in Fig.~\ref{fig:rantResults}$(g)$).

In order to quantify the robustness of the construction algorithm, we vary the strength of phototaxis (captured via $\C$) and the threshold (captured via $c^*$). In Fig.~\ref{fig:robust}$(a)$ and SI Fig.~\clusterNumber$(b)$ we show the results of these manipulations. With no threshold i.e. $c^*$ = 0, obstacles are randomly dropped in the construction area and in particular around the boundary (see Fig.~\ref{fig:robust}$(a)$). In the presence of a detachment threshold $c^*$ but without phototaxis $\C=0$, clusters are formed but the process is very slow.  When both the detachment threshold, $c^*$ and the phototaxis mechanism is active $\C=1$ we observe a coherent final structure. The circumference length $L(\lambda_\text{a}, \lambda_\text{b})$ (shown in SI Fig.~\clusterNumber$(b)$) and the relative area coverage of the largest cluster $A_\text{e}/A_\text{o}$ capture the robustness of the coupled mechanism when the detachment threshold $c^*$ and phototaxis $\C$ are active or inactive. For an inactive threshold corresponding to $c^*=0$, we generally see elongated ellipses and small clusters (see SI Fig.~\clusterNumber$(b)$) that indicate a low cooperative effort and random substrate element deposition. For a case with active threshold $c^* \neq 0$ but no gradient information $\C=0$ we do see nearly circular clusters which indicate a localized construction effort, with small clusters indicating a cohesive (but slow) construction process. Only for active threshold $c^* \neq 0$ and gradient information $\C=1$  do we observe fast, localized  and large constructions.\\

\begin{figure*}[t!]
\centering
\includegraphics[width=0.9\textwidth]{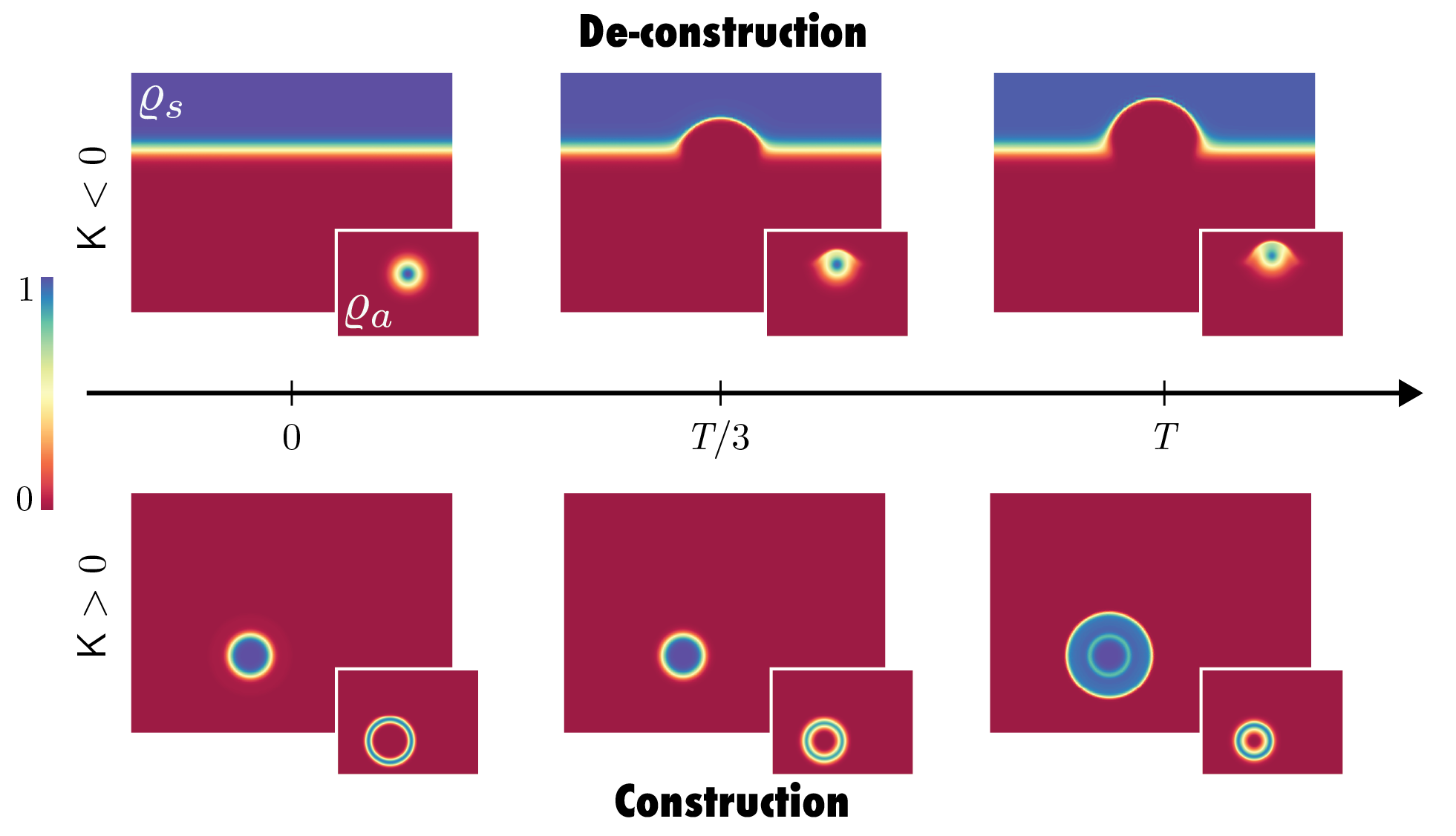}
\caption{\textbf{\textsf{Continuum simulations}} Simulations of the Eqs.~\ref{eq:rhor}-\ref{eq:rhos} showing substrate density $\varrho_s$ (and density $\varrho_a$ in insets) capturing both cooperative de-construction for $\T = -1.44$ and construction for $\T = 1.44$ and $\C = 0.8$ (see sec.~\simDet for simulation details). The agent density, $\varrho_a$ propagates into the interior of the substrate resulting in degradation of the density $\varrho_s$ for $\T < 0$ while the substrate density $\varrho_s$ grows in magnitude in locations where the agents cluster for $\T > 0$. This captures the tasks performed by the RAnt collective defined by the cooperation parameter, $\C$ vs the deposition rate $\T$ into a unified phase space. Color bar indicates the contour values of $\varrho_a, \varrho_s$.}
\label{fig:2Dsimul}
\end{figure*}

\noindent \textit{Continuum model of collective architectonics:} So far we have seen the efficacy and robustness of the chosen interaction rules in the triadic model for collective construction in a discrete setting (shown in Fig.~\ref{fig:intro}$(d)$). We now capture dynamics of the collective by averaging over the microscopic behaviors (corresponding to small time-scale processes) of the RAnts and arrive at a continuum model to understand the macroscopic behavior of the collective~\cite{Prasath2021cooperative}. Our model for collective construction involves three interacting spatio-temporal fields which are the agent density $\varrho_a (\x, t)$, photormone field  $c (\x, t)$ and substrate density $\varrho_s (\x, t)$ (shown schematically in Fig.~\ref{fig:intro}$(d)$). Our model for the dynamics of cooperative task execution extends the Patlak-Keller-Segel model~\cite{hillen2009user,Prasath2021cooperative,prasath2022rheomergy} for aggregation in biological systems by accounting for how the architectural environment changes and in-turn modulates RAnt behavior. Here the agent density $\varrho_a (\x, t)$ represents the averaged position of the RAnts averaged over a characteristic time-scale set by the time taken to traverse several domain lengths. The agent density travels with a self-propelled velocity $\u_a$ which is related to the local environment through the relation $\u_a = v_o (1-\varrho_s) \hat{\mathbf{p}}$ where $v_o$ is the speed of the collective and $\hat{\mathbf{p}}$ is its orientation. Just as in the case of RAnts, the velocity of the agents is modified by gradients in photormone concentration due to phototaxis and is given by $\chi \grad c$ where $\chi$ is the phototaxis strength. In the absence of such gradients, since RAnts use an exploratory strategy, their density can diffuse out with diffusivity $D_a$. We generalize the photormone dynamics in Eq.~\ref{eq:photoDless}, with $k_+, k_-$ as the rate of production and decay of photormone, by also accounting for its diffusivity, $D_c$ (just as seen in pheromones). The construction process is defined as the transportation of substrate elements to locations of high photormone concentration beyond a critical threshold in concentration at a particular location. The dynamics of the substrate density $\varrho_s(\x, t)$ in the construction process is captured using a thresholded rate equation where the construction rate is given by $k_s$.\\

The dynamical equations for $\varrho_a (\x, t)$, $c (\x, t)$ and $\varrho_s (\x, t)$ becomes (see SI sec.~\contmodel for details)
\begin{align}
\partial_t \varrho_a =& -\underbrace{\nabla \cdot (\u_a \varrho_a)}_{\substack{\text{Self-propulsive} \\ \text{advection}}}  + \ \nabla \cdot (\underbrace{D_a \grad \varrho_a}_{\text{Diffusive flux}} - \underbrace{\chi \varrho_a \grad c}_{\text{Phototaxis}}), \label{eq:rhor}
\end{align}

\begin{align}
\partial_t c =& \ \underbrace{D_c \nabla^2 c}_{\text{Diffusion}}\ + \underbrace{k_+ \varrho_a}_{\text{Production}} -\ \underbrace{k_- c}_{\text{Decay}}, \label{eq:cp}
\end{align}

\begin{align}
\partial_t \varrho_s =&\ k_s \varrho_s \{ \ \Theta \underbrace{(c-c^*)}_{\substack{\text{Photormone} \\ \text{field threshold}}} \ \} \times \{ \ \Theta  \underbrace{ (\varrho_a-\varrho^*_a)}_{\substack{\text{Robot density} \\ \text{threshold}}}\ \}.  \label{eq:rhos}
\end{align}
In Eq.~\ref{eq:rhos}, $\varrho^*_a$ and $c^*$ are the threshold concentration of agent density and photormone field required to initiate construction. The threshold behavior that determines the consensus of the agents to start construction is captured via the Heaviside function $\Theta(x)$ in Eq.~\ref{eq:rhos}. Just as in the case of discrete agents in Eqs.~\ref{eq:pos},~\ref{eq:orn}, the cooperation parameter is quantified as the relative strength of phototaxis to diffusion: $\C = \chi c_o/D_a$ with $c_o$ being the reference photormone concentration (discussed in SI sec.~\contmodel). The other relevant non-dimensional number capturing the task is the non-dimensional deposition rate $\T = k_s l/v_o$ where $l$ is the characteristic length-scale obtained from photormone diffusion-decay dynamics, $l \sim (D_c/k_-)^{1/2}$ (SI sec.~\contmodel for further details). We note that in our continuum model, $\C \in [0, \infty), \K \in (-\infty, \infty)$, in contrast to the discrete case where $\C \in [0,1], \K \in [-1,1]$, but both the models capture the same qualitative features. We see in Fig.~\ref{fig:2Dsimul} that for $\T > 0$ the collective performs construction and whereas for $\T < 0$ they perform de-construction, i.e. a simple flip in sign of the deposition rate (associated with agent-environment interaction) results in macroscopically antagonistic tasks. This is indeed what we observe in experiments (shown in Fig.~\ref{fig:robust}$(b)$) where the RAnts cooperatively tunnel through a bulk substrate when $\K<0$, and cooperatively construct when $\K>0$ (ref. Fig.~\ref{fig:2Dsimul}). Together, the cooperation parameter $\C$ and the deposition rate $\T$ span a two dimensional phase space of macroscopic behaviors with their boundaries separated by parametric instability. This allows us to choose from a selection of behaviors that range from coordinated construction to scattered construction and collective excavation on the fly, by simply tuning these parameters.

\noindent \textit{Conclusions:} Biologically inspired collective robotics allows us to explore the emergence of complex architectonics from simple individual behaviors that are coupled to spatio-temporal environmental fields. In our synthetic system, the robotic agents leverage a spontaneous trapping instability to reach a concensus and nucleate a construction site that is then cooperatively expanded by phototaxis. Furthermore, the macroscopic behaviors exhibited by the collective goes beyond cooperative construction. By a simple change in the behavioral rule of the RAnts we see that the agents can change from construction to de-construction behavior robustly, consistent with our continuum theoretical framework. 

The phase-space of functional behavior shown here is a first step towards understanding and synthesizing adaptable collective systems where the agents cooperatively change their behavior to perform the necessary task on demand. Perhaps the collective construction observed in social insects emerges from a process similar to the mechanism presented here, and will bring us a step closer to understanding and synthesizing the plethora of functional collective behaviors observed in the natural world.

\begin{acknowledgments}
{We thank the NSF PHY1606895 (S.G.P., L.M.), Swiss National Science foundation P400P2-191115, NSF EFRI 18-30901 (L.M.), NSF 1764269 (L.M.), Kavli Institute for Bionano Science and Technology (L.M.), the Simons Foundation (L.M.) and the Henri Seydoux Fund (L.M.) for partial financial support.} 
\end{acknowledgments}

\bibliographystyle{unsrtnat}
\bibliography{references}

	\widetext
	\clearpage
	\onecolumngrid
	\begin{center}
	\textbf{\large Supplemental Materials: Collective phototactic robotectonics}\\[.2cm]
	Fabio Giardina,$^{1, *}$ S  Ganga  Prasath,$^{1, *}$ and  L  Mahadevan$^{1, 2, 3, 4}$\\[.1cm]
	{\small \itshape ${}^1$School of Engineering and Applied Sciences, Harvard University, Cambridge MA 02138.\\
	${}^2$Center for Brain Science, Harvard University, Cambridge, MA 02138.\\
	${}^3$Department of Physics, Harvard University, Cambridge MA 02138.\\
	${}^4$Department of Organismic and Evolutionary Biology, Harvard University, Cambridge 02138.\\
	${}^*$equal contribution}
	\end{center}
	\setcounter{equation}{0}
	\setcounter{figure}{0}
	\setcounter{table}{0}
	\setcounter{page}{1}
	\makeatletter
	\renewcommand{\theequation}{S\arabic{equation}}
	\renewcommand{\thefigure}{S\arabic{figure}}
	\renewcommand{\bibnumfmt}[1]{[#1]}
	\renewcommand{\citenumfont}[1]{#1}
	\linespread{2.5}

	\section{SI videos}
	\begin{itemize}
	\item \href{https://www.dropbox.com/s/2awy4ed8v2vwsbz/SIVid_1.mp4?dl=0}{\underline{Video 1}}: Single RAnt moving in a circular trajectory under fixed projected field with constant gradient in photormone intensity.
	\item \href{https://www.dropbox.com/s/4myr4f6tefc1zr5/SIVid_2.mp4?dl=0}{\underline{Video 2}}: Single RAnt and two RAnts trapping in their self-generated photormone field.
	\item \href{https://www.dropbox.com/s/81mwu673t5nvf8k/SIVid_3.mp4?dl=0}{\underline{Video 3}}: Cooperative construction for $\C = 1$ where RAnts nucleate a construction site via trapping instability and use it to construct a structure. Also shown is incoherent construction when $\C = 0.25$ where the clusters constructed are dispersed. Construction elements outside the construction area are being manually rearranged during the experiment to ensure an even distribution of construction material at the arena boundaries.
	\item \href{https://www.dropbox.com/s/ibhea5jigk4zn4m/SIVid_4.mp4?dl=0}{\underline{Video 4}}: Collective de-construction where RAnts excavate a location which has higher photormone concentration.
	\end{itemize}

All supplementary videos can be accessed either directly by clicking on their respective link or \href{https://www.dropbox.com/sh/9tl7r6scep68sxd/AAC-W1AzZCgntow7mh0v2rO5a?dl=0}{\underline{here}}.

\section{Single RAnt dynamics}\label{sec:SISRAnt}
The governing equation for the position and orientation of the RAnts is given by,
\begin{align}
\dot{\r} =& \ v_o \ph,  \label{eq:SIpos} \\
\dot{\ph} =& \ \Omega \times \ph,  \label{eq:SIorn}
\end{align}
where the RAnt is programmed such that $\Omega = G(\grad c \cdot \nh) \hat{z}$.

\subsection{Constant gradient}
\begin{figure*}[t!]
\centering
\includegraphics[width=\textwidth]{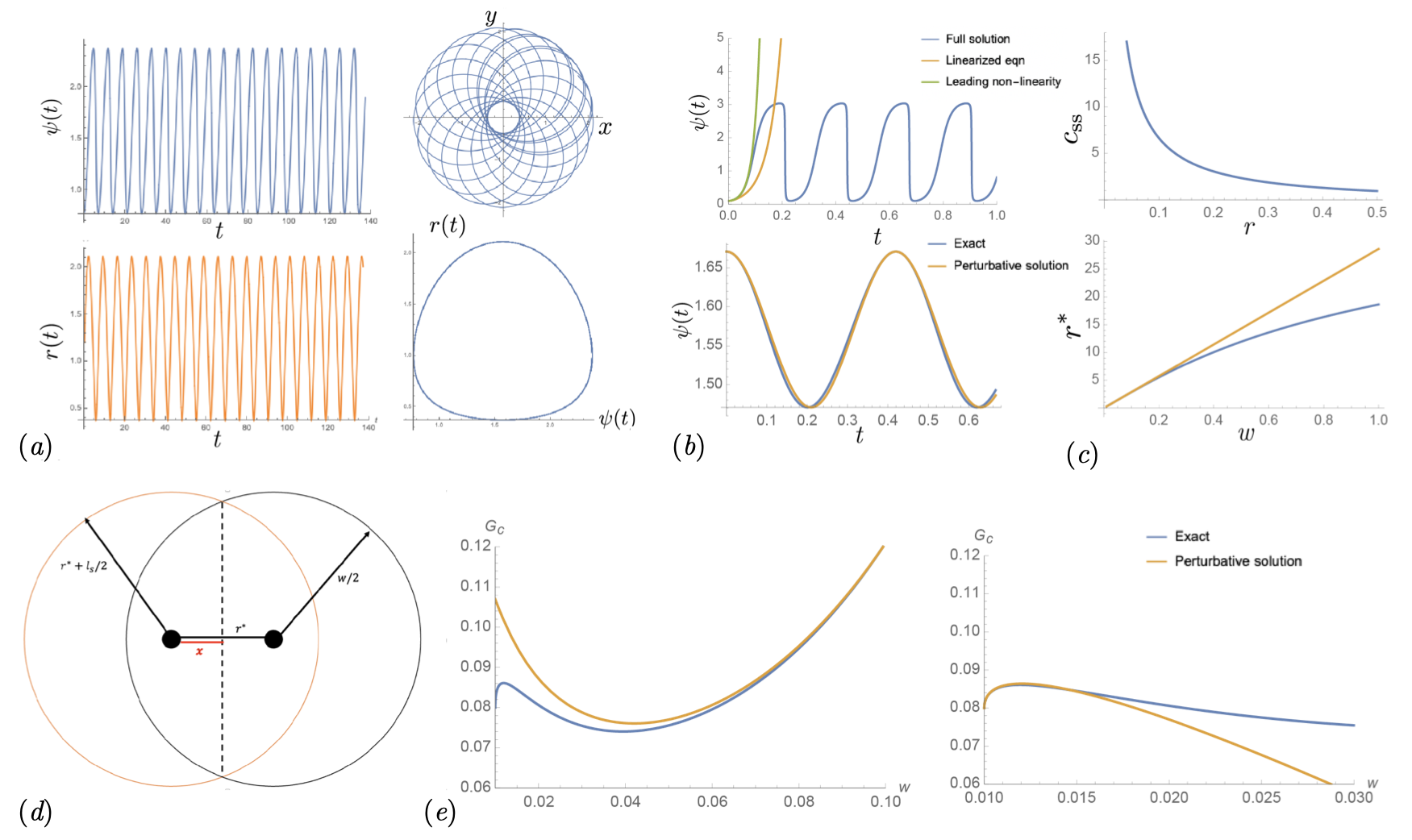}
\caption{$(a)$ Evolution of $\psi(t), r(t)$ and the corresponding evolution in $(x,y)$-coordinates, $(r, \psi)$-plane from Eqs.~\ref{eq:SIr},~\ref{eq:SIpsi} with $r(0) = 1.2, \psi(0) = \pi/4, \gt = 1$. We find periodic dynamics in $(r, \psi)$-plane. $(b)$ (\textit{Top}) Evolution of $\psi$ from Eq.~\ref{eq:SIpsiFull} and comparison with linearized solution valid for short times and leading order non-linear solution. We set $\psi(0) = \dot{\psi(0)} = 0.1$. (\textit{Bottom}) Comparison of solution $\psi(t)$ with perturbative solution obtained in Eq.~\ref{eq:SIpsiPer} for $\gt = 15$ and initial conditions $\psi(0) = (\pi/2+0.1), r(0) = r^*$. $(c)$ Solution to photormone profile, $c_{\text{ss}}$ vs $r$ when $l_- \gg w$ in Eq.~\ref{eq:SIcss} and comparison of $r^*$ obtained from the steady state profile with leading order linear behavior. $(d)$ Schematic showing the radius $r^*$ at which RAnts undergo periodic motion and other relevant variables. $(e)$ Comparison of $G_c$ vs $w$ with in the two limits derived in Eqs.~\ref{eq:SIgcsmall},~\ref{eq:SIgclarge} when $l_{\s} = 0.01, k_+ = 1.5, k_- = 1.5, v_o = 0.04$.}
\label{fig:SIFig1}
\end{figure*}

In the presence of a constant gradient such that $\grad c = -\lambda \rh$ where $\rh = (\cos \phi, \sin \phi)$, $\ph = (\cos \theta, \sin \theta), \nh = (-\sin \theta, \cos \theta)$. From this we have that $\Omega = G \lambda \sin(\theta-\phi) \hat{z}$. This implies that the rotation dynamics is $$\dot{\ph}= G\lambda \sin(\theta-\phi) \nh.$$ We now move to complex notation where $\r(t) = r(t) \e^{i \phi(t)}, \ph = \e^{i\theta}, \nh = \e^{i (\theta+\pi/2)}$. From this we have, $\dot{\theta} = \lambda \sin (\theta - \phi)$. Further the position of the RAnt is: $[\dot{r} + i r \dot{\phi}]\e^{i \phi} = \ v_o \e^{i\theta}$. Ultimately combining the equations gives,
\begin{align}
\dot{r} =& \ v_o \cos (\theta - \phi), \\
\dot{\phi} =& \ \frac{v_o}{r} \sin(\theta - \phi), \\
\dot{\theta} =& \ G \lambda \sin (\theta - \phi).
\end{align}
These equations can be rewritten by redefining $\psi = (\theta - \phi)$ and we get,
\begin{align}
\dot{r} =& \ v_o \cos \psi, \\
\dot{\psi} =& \ \bigg( G\lambda - \frac{v_o}{r} \bigg) \sin \psi.
\end{align}
This can be written in non-dimensional form by using $\lambda^{-1}$ as the length scale and $1/(v_o \lambda)$ as the time scale as
\begin{align}
\dot{r} =& \ \cos \psi, \label{eq:SIr} \\
\dot{\psi} =& \ \bigg( \gt - \frac{1}{r} \bigg) \sin \psi. \label{eq:SIpsi}
\end{align}
The fixed points of this system is $\psi = \pi/2, r = 1/\gt$ where $\gt = G/v_o$. The dynamics of this system can show periodic and quasi-periodic dynamics as shown in Fig.~\ref{fig:SIFig1}$(a)$. This system can also be written as a second order system as
\begin{align}
\ddot{\psi} =& \ \gt^2 \cos \psi \sin \psi  +
2 \dot{\psi} \cot \psi [\dot{\psi} - \gt \sin \psi]. \label{eq:SIpsiFull}
\end{align}
Linearizing this equation we get $\ddot{\psi} = \gt^2\psi - 2\gt \dot{\psi}$ and this can be solved to get:
\[
\psi(t) = \frac{\e^{-\gt t}}{2\gt} \bigg[ \sqrt{2} (a+b \gt) \sinh \left(\sqrt{2} \gt t\right) + 2 b \gt \cosh \left(\sqrt{2} \gt t\right) \bigg],
\]
where $\psi(0) = a, \dot{\psi}(0) = b$. For small amplitudes i.e. $\psi \ll 1$, we can obtain the solution to leading order non-linear equation: $\ddot{\psi} = \gt^2 \psi - 2 \gt \dot{\psi} + 2(\dot{\psi})^2/\psi$. The solution to this equation is $\psi(t) = \dfrac{a^2 \e^{\gt t}}{a - b t + a \gt t}$. We compare the evolution of both the solutions with the full solution in Fig.~\ref{fig:SIFig1}$(b)$. Now in order to understand the periodic dynamics close to the fixed point, $r^* = 1/\gt, \psi = \pi/2$, we perturb the variables: $\psi = (\pi/2 + \pst), r = r^* + \rt$ where $\pst \ll \pi/2, \rt \ll r^*$. The evolution equation then becomes
\begin{align}
\ddot{\pst} +\gt^2 \cos \pst \sin \pst + \dot{\pst}^2 \tan \psi =& \ 0.
\end{align}
When $\gt \gg 1$, this equation becomes $\ddot{\pst} + \gt^2 /2 \sin 2\pst \approx 0$. The solution to this will give us the dynamics close to the fixed points. In order to find the period of oscillation of $\pst(t)$ we can write the solution after using Poincar\'e-Lindstedt method to obtain
\begin{equation}
\pst(t) = 2 \alpha \cos \bigg(\beta - \frac{\alpha^2}{\gt^2} t + \gt t \bigg), \label{eq:SIpsiPer}
\end{equation}
where the constants $\alpha, \beta$ and set by the boundary conditions $\pst(0), \dot{\pst}(0)$. We compare the full solution with this solution in Fig.~\ref{fig:SIFig1}$(b)$.

\subsection{Dynamic photormone field}\label{sec:dynamic}

The time evolution of the photormone field is given by
\begin{align}
\partial_t c(\x,t) =& \ k_+ \varrho_a[w, l_{\s}] - k_- c(\x,t),\label{eq:SIc}
\end{align}
where $\varrho_a$ is the region around the RAnt location over which photormone is produced. As we have discussed in the main text tn the coupled dynamics of the RAnts (Eqs.~\ref{eq:SIpos},~\ref{eq:SIorn}) and photormone, there are 3 intrinsic length-scales: $w, l_{\s}, l_- (= v_o/k_-)$ where $w$ is the width of the region over which photormone is produced and $l_{\s}$ is the distance between the sensors and lastly the length scale over which the RAnts travel before the photormone they generate decays $v_o/k_-$. Since steady states are given by RAnts reaching a fixed $r^*$, this is given by the fixed points found earlier for a fixed gradient. We perform analysis in two limits of scale-separation where the RAnts exhibit trapped dynamics given by $l_- \ll w \sim l_{\s}$ and $w \sim l_{\s} \ll l_-$. We can write these limits using non-dimensional parameters as $\L_- \ll \L_w \sim \O(1)$, $\L_w \sim \O(1) \ll \L_-$ where $\L_- = l_-/l_\s, \L_w = w/l_\s$. When the length-scale over which pheromone decays is small, the gradient measured by the sensor is determined by the size over which they generate and is given by $w$. In this limit the RAnts choose a radius such that they can measure maximum gradient and this happens at a radius of $r^* = (w +  l_{\s})/4$ which is the trapping radius. We can write this in non-dimensional terms as $\ndr^* = (\L_w + 1)/4$. Though this is the radius of trapping, one must ensure that the RAnts are capable of making this turning radius. For a given gradient in photormone field we know that the critical turning radius is given by $\ndr_c = 1/(\gt \grad c)$.

In order for the trapping radius $\ndr^*$ to be valid, we require that it be greater than the critical turning radius $\ndr_c$ i.e. $\ndr^* > \ndr_c$. We can identify the boundary of self-trapping using the conditions $\ndr^*=\ndr_c$. The unknown in this equation however is $\grad c$. For the given $\ndr^*$, the sensor closer to the center of the trap is bound to measure a constant photormone field $c=k_+/k_-=\kh_\pm$ since the photormone production diameter $w$ is large enough to cover the circle drawn by the trajectory of that sensor or effectively $\L_w \gtrsim 1$. Hence the left sensor value is governed by the equation $\dot{c} = \kh_\pm - c$ and the outer sensor travelling on the circle with radius $\R=(\ndr^*+1)/2$ ``sees'' photormone production only for a fraction (which is unknown) of the inner sensor. It is worth reminding here that these equations have been non-dimensionalized using $(1/k_-)$ as the time-scale and $l_s$ as the length-scale. Therefore, $\grad c = [ c(\R) - \kh_\pm]$ and the value of photormone at the outer sensor converges to
\[
c(\R) = \kh_\pm \bigg[ 1-\exp \bigg( \frac{\tau_1}{2} \bigg) \frac{\exp(\tau_2) - 1}{\exp(\tau_1 + \tau_2) - 1} \bigg],
\]
with $\tau_1 + \tau_2 = 2\pi \ndr^*/\L_-$. The position of intersection of the two circles (ref. Fig.~\ref{fig:SIFig1}$(d)$) can be written in dimensional form as $x = ({r^*}^2-(w/2)^2+R^2)/2 r^*$ with $R=(r^*+l_{\s})/2$ which equivalently becomes $\mathsf{x} = ({\ndr^*}^2-(\L_w/2)^2+\R^2)/2 \ndr^*$ in non-dimensional form. The angle between the horizontal and the upper intersection point is $\vartheta = \arccos (\mathsf{x}/\R)$ and the arc length is therefore $2\R\vartheta = 1$. Thus, the ratio is $\rho = \vartheta/\pi = \arccos (\mathsf{x}/\R)/\pi$. With $\mathsf{x}/\R = (5-\L_w)/(3+\L_w)$, we get $\tau_1=(2\ndr^*\arccos(\mathsf{x}/\R))/\L_-$. From this we can write the full expression becomes $\gt_c = 1/[\kh_\pm - c(\R)]$. When $\L_w \sim \O(1)$, we can expand the variables $\rho = \sqrt{\epsilon}/\pi$ where we have used $\L_w = (1+\epsilon)$. Further we get $\tau_1 = \sqrt{\epsilon} / \L_-, \tau_2 = \pi/\L_- - \sqrt{\epsilon}/\L_-$.
Equating the expression for $\ndr_c$ with $\ndr^* = ( 1 + \epsilon/2)/2$, we find that (shown in Fig.~\ref{fig:SIFig1}$(e)$)
\begin{equation}
\gt_c = \frac{2}{\kh_\pm} + \frac{\L_w \sqrt{\epsilon}}{\L_- \kh_\pm} \coth \bigg( \frac{ \pi \L_w}{2 \L_-} \bigg), \label{eq:SIgcsmall}
\end{equation}
On the other limit when $\L_w \gg 1$, we find that (also shown in Fig.~\ref{fig:SIFig1}$(e)$)
\begin{equation}
\gt_c = \frac{4 \L_-}{\L_w \kh_\pm} \sinh \bigg( \frac{\pi \L_w}{4 \L_-} \bigg).
\label{eq:SIgclarge}
\end{equation}

\begin{figure}[t!]
\centering
\includegraphics[width=0.85\textwidth]{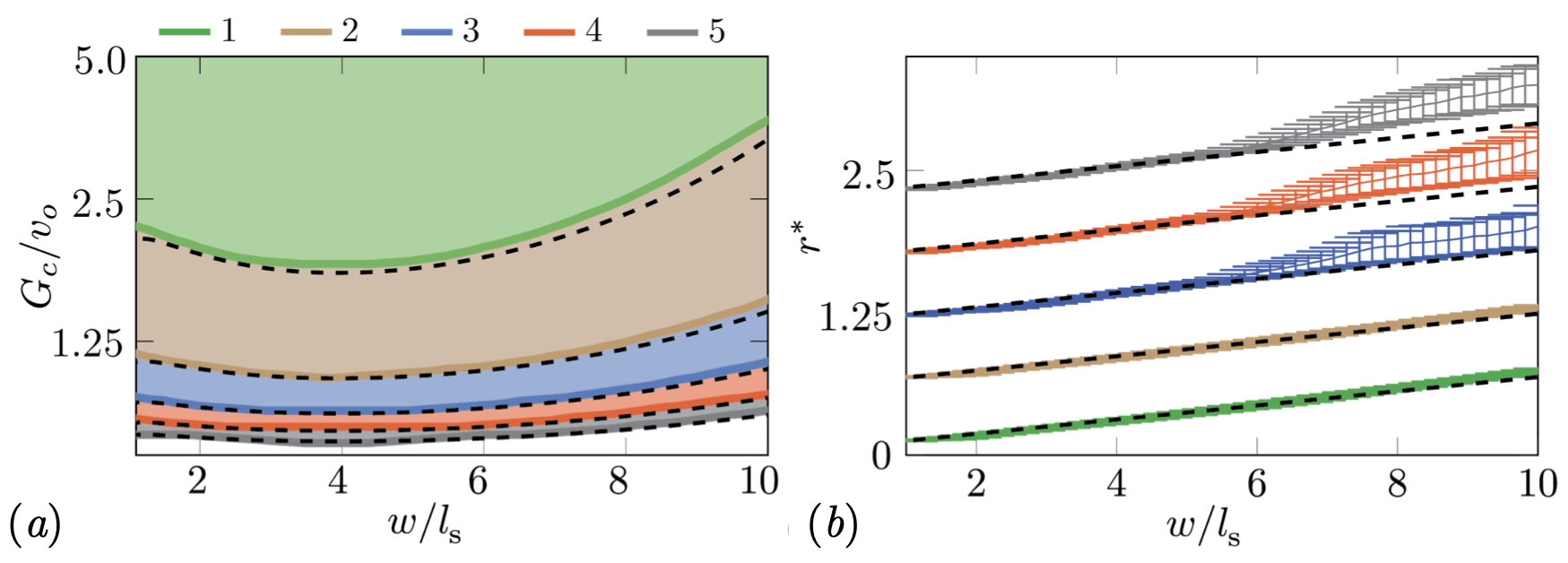}
\caption{Trapping mechanism for RAnts capturing using non-dimensional gain of each RAnt vs the production width of photormone. $(a)$ Theoretical predictions for boundaries of trapping and $(b)$ radii of trapping for different number of RAnts (1-5) are shown as dashed lines and the simulations are the filled areas trapping parameters. Note that the trapping radius is independent of the number of RAnts and the individual affine functions are offset vertically as they would overlap otherwise.}
\label{fig:SIFigTrap}
\end{figure}

When $\L_-$ is large enough, the geometry associated with the generation of photormone no more affects the trapping dynamics as $\L_w \ll \L_-$. We can assume that a source of constant width and write the evolution of the concentration field at a location $\ndr$ where the RAnt has traversed based on the duration spent by the RAnt at that location. The evolution equation when the RAnt is location at a location $\ndr$ is given by: $\dot{c}(t) + c(t) = \kh_\pm$ for $t \in [\tau^a_i, \tau_i + \L_w/\L_-]$ where $\tau^a_i$ is the time at which the top part of the RAnt reaches $(\ndr, \phi)$ which can be written without loss of generality as $(2\pi i \ndr/\L_-), i \in \Z$ if the RAnt revolves at a distance $r$ from the origin. When the RAnt is not located at a point, the concentration evolves as $\dot{c}(t) + c(t) = 0$ for $t \in [\tau^a_i + \L_w/\L_-, \tau^a_{i+1}]$. Steady state concentration is reached when the peak amplitude does not change. We can obtain the solution for the concentration during the active phase as
\[
c(t) = \kh_\pm + ( c_o - \kh_\pm ) \e^{-t},
\]
and during the decay phase as
\[
c(t) = c_1 \e^{-t}.
\]
A steady state is reached when the peak concentration produced by the RAnts during the active phase does not change. Under this setting, we can obtain the peak concentration as (shown in Fig.~\ref{fig:SIFig1}$(c)$)
\begin{equation}
c_\ss(\ndr) = \frac{\kh_\pm}{2} \left(\e^{\L_w/\L_-}-1\right) \left[ \coth \left(\frac{\pi \ndr}{\L_-}\right)-1 \right]. \label{eq:SIcss}
\end{equation}
We know that a steady state in RAnt motion is reached when $\ndr^* = 1/(\gt \grad c)$. Upon substituting we get an implicit equation
\[
\ndr^* = \frac{2 \L_-}{\pi \gt \kh_\pm} \sinh^2 \left(\frac{\pi \ndr^*}{\L_-}\right) \frac{1}{\e^{\L_w/\L_-} - 1}.
\]
From this we immediately see that for $\L_- \gg \L_w, \ndr^*$ we get $\ndr^* \approx \L_w \gt \kh_\pm/2\pi$. Figure~\ref{fig:SIFig1}$(c)$ shows the comparison between the full solution and the leading order linear behavior.

Figure~\ref{fig:SIFigTrap}$(a)$ shows theoretical predictions in Eqs.~\ref{eq:SIgcsmall},~\ref{eq:SIgclarge} (dashed lines) and simulations (shaded areas, see next section) of regions of trapping. A single RAnt traps in the green shaded region, while two agents trap in the brown region (and above), three in the blue region (and above), and so forth. For multiple RAnts, the critical trapping gain $\gt_c$ scales inversely with the number of RAnts, $\gt_c \sim 1/n$ (where $n$ is the number of RAnts). We confirm this in simulations as shown in Figure~\ref{fig:SIFigTrap}$(a)$. Trapping stops a RAnt from exploring the space it operates in and causes a build-up of photormone in a confined region of space. We can avoid trapping of a single agent by operating below the green region and thus have control over the number of agents that must be within the trapping radius (see Fig.~\ref{fig:SIFigTrap}$(b)$) to start a trap.

\section{Simulation of RAnt dynamics}
In order to understand the dynamics of RAnts in the presence of a dynamically changing photormone field, we couple Eqs.~\ref{eq:SIpos}-\ref{eq:SIorn} with Eq.~\ref{eq:SIc} and solve them numerically. We use \textsc{Matlab} to evolve this dynamical system for RAnts and photormone where the RAnt dynamics is solved using a fourth-order Runge-Kutta integration scheme with adaptive step size while the photormone field is evolved by discretizing the space and time marching using a second-order accurate finite difference scheme. The agent density field computed from the position of the RAnt is assumed to be Gaussian of width $w$. We initialize RAnts randomly in a photormone field that has a linear gradient that decays with time. This gradient helps in introducing rotation motion of the RAnts in the simulation. There is no hard repulsion between the RAnts in the simulation and they can pass through each other and the photormone field is the only medium of communication between the RAnts. In experiments, however, the RAnts have a bump sensor through which they can also interact with each other. The parameters involved in the simulation are the RAnt's width $w$, rotation gain $G$, speed $v_o$, sensor distance $l_{\s}$, photormone production and decay rates $k_+, k_-$. The simuation domain is $L$ and we assume periodic boundary conditions on the sides of the domain.

\begin{figure*}[t!]
\centering
\includegraphics[width=0.9\textwidth]{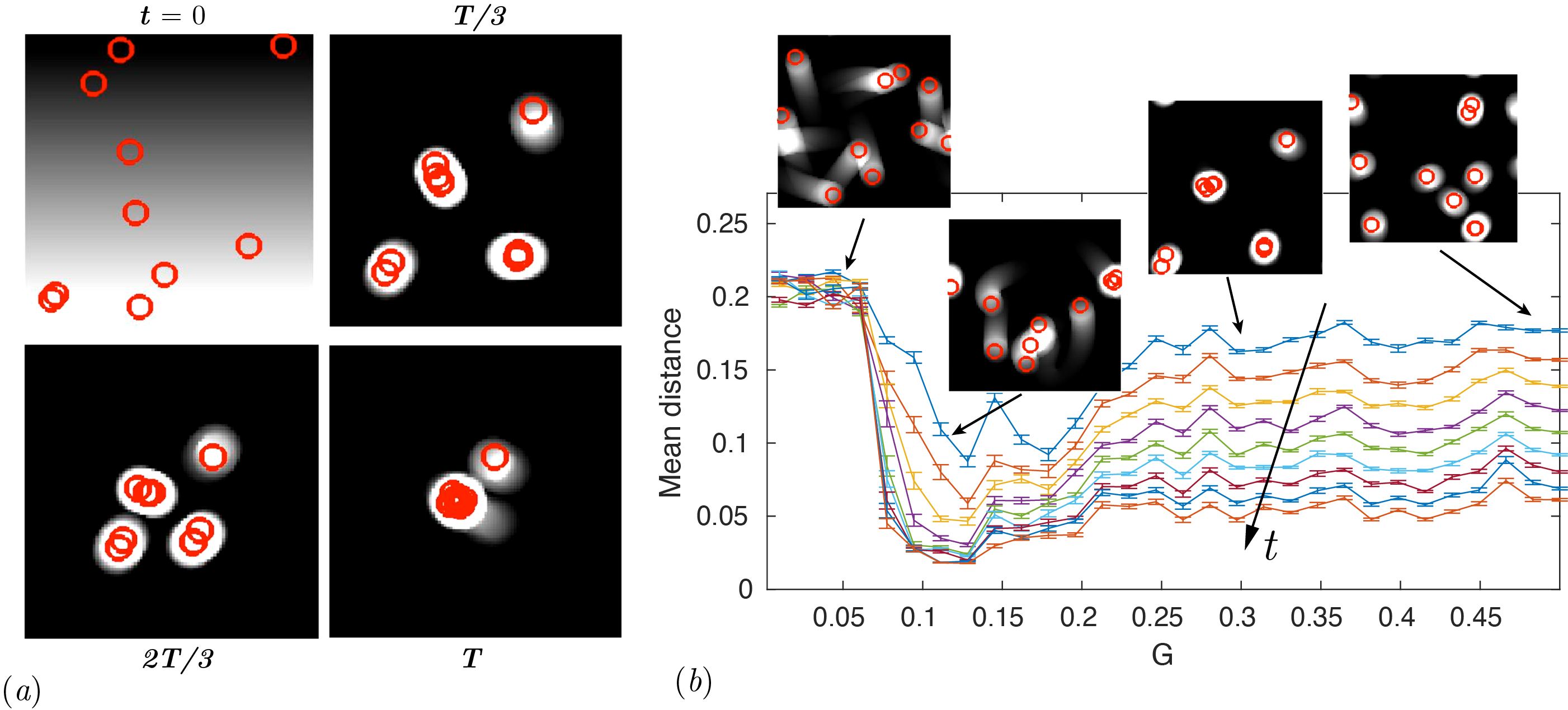}
\caption{$(a)$ Snapshots of photormone field with the RAnt locations at 4 different time instants from numerical simulations when the width of photormone generated $w = 0.03$, rotation gain $G = 0.3$, speed $v_o = 0.04$, sensor distance $l_{\s} = 0.01$, $k_+ = k_- = 1.5$, number of RAnts $n_r = 10$, domain length $L = 0.2$. $(b)$ The mean distance between RAnts for different values of $G$ averaged over 30 simulations. We see that for large values of $G$, the mean distance between traps reduces and eventually there exists only one trap.}
\label{fig:SIFig2}
\end{figure*}

We find that for small values of $G$, RAnts do not get trapped and keep moving inside the domain resulting in large average distance between them at all times (see Fig.~\ref{fig:SIFig2}$(a, b)$). As we increase the rotation gain, there is a critical value at which RAnts get trapped (as we have discussed in detail earlier). The average distance between the RAnts is still large but because of small perturbations we see that the RAnts exhibit quasi-periodic dynamics that keeps the trap in motion. This process results in coarsening of the traps where initially large number of self-trapping happens for large values of $G$ which later combine to result in a single trap, which is captured through mean distance between RAnts as a function of time as shown in Fig.~\ref{fig:SIFig2}$(b)$.

\section{Continuum model}\label{sec:contModel}
As we have described in the main text the 3 relevant variables required to capture cooperative task execution are the robot-density $\varrho_a({\bf x},t)$, photormone field $c({\bf x},t)$ and the substrate density $\varrho_s({\bf x},t)$. The dynamical equations in dimensional form their evolution are given by,
\begin{align}
\partial_t \varrho_a + \nabla \cdot (\u_a \varrho_a) =& \ \nabla \cdot (D_a \grad \varrho_a - \chi \varrho_a \grad c), \label{eq:SIrhor}\\
\partial_t c =& \ D_c \nabla^2 c + k_+ \varrho_a - k_- c, \label{eq:SIcp} \\
\partial_t \varrho_s =& \frac{1}{4} k_s \varrho_s (1 + \tanh [\alpha_c (c-c^*)]) \nonumber \\
& (1+\tanh [\alpha_c (\varrho_a-\varrho^*_a)]). \label{eq:SIrhos}
\end{align}
This model has been analyzed in detail in the de-construction phase in~\cite{Prasath2021cooperative}. We regularize the Heaviside function, $\Theta(x)$ in Eq.~\rhos using the hyperbolic function $[1+\tanh(x)]/2$ and the velocity of the group is $\u_a = v_o(1-\varrho_s/\varrho_o)\nh$ captures the decrease in the magnitude of the speed due to collision with the substrate elements. There are a total of 7 time-scales associated with different processes that the model captures which are listed in tab.~\ref{tab:timeScale}.

From the 7 time-scales in the model we can construct the following 5 non-dimensional numbers that capture the dynamics of the collective,
$$
\C = \frac{\chi c_o}{D_a}, \ \T = \frac{k_s l}{v_o}, \ \vh  = \frac{v_o l}{D_a}, \ \kh_\pm = \frac{k_+ \varrho_o}{k_- c_o}, \ {\dh}_c = \frac{D_c}{l^2 k_-}.
$$
We can rewrite the Eqs.~\ref{eq:SIrhor}-\ref{eq:SIrhos} in non-dimensional form as
\begin{align}
\partial_t \varrho_a +\nabla \cdot [ (\C \grad c + \vh (1-\varrho_s)) \varrho_a ] =& \ \grad^2 \varrho_a , \label{eq:SIndrhoa}\\
\partial_t c =& \ \dh_c \nabla^2 c + \kh_\pm \varrho_a - c, \label{eq:SIndc} \\
\partial_t \varrho_s =& \frac{1}{4} \T \varrho_s (1 + \tanh [\alpha_c (c-c^*)]) \times \nonumber \\
& (1+\tanh [\alpha_c (\varrho_a-\varrho^*_a)]). \label{eq:SIndrhos}
\end{align}

\subsection{Solution to photormone dynamics}
The photormone dynamics given by Eq.~\ref{eq:SIndc} has 3 time-scales associated with production, diffusion and decay. In the diffusion dominated limit, the steady-state solution to the Poisson equation is given by:
\[
c(r) = \frac{\kh_\pm }{4 \pi \dh_c} \int_S \frac{\varrho_a (r')}{|r-r'|} \ \d r',
\]
whereas in the decay dominated limit we have $c(r) = \kh_\pm \varrho_a(r)$.

The photormone dynamics is a linear equation in photormone concentration. This equation can be solved exactly using the method of Green's function. Before we find the Green's function, we make the transformation $c = \e^{-t} u$, then we can write: $\partial_t c = \e^{-t}\partial_t u - c$. Substituting this in the photormone dynamics, we get
$$
\partial_t u = \dh_c \nabla^2 u + \kh_\pm \e^{t} \varrho_a.
$$
The Green's function for the heat equation with inhomogeneous time-dependent forcing can be found by solving the system, $\partial_t g = \dh_c \nabla^2 g + \delta(t-\tau) \delta (\x - \x')$. After solving this, the solution can be ultimately written as $ u(\x, t) = k_+  \int_{0}^\infty \int_V g(\x, t; \x', \tau) \e^{\tau} \varrho_a \ \d \x \ \d \tau$. This can be solved to get the Green's function as: $g(\x,t; \x', \tau) = \frac{ \Theta (t - \tau)}{4 \pi \dh_c (t-\tau)} \exp \bigg({-\frac{|\x - \x'|^2}{ 4 D_c (t-\tau)}} \bigg)$. The ultimate solution to the transformed equation is,
$$
u(\x, t) = \kh_\pm  \int_{0}^\infty \int_V \frac{ \Theta (t - \tau)}{4 \pi D_c (t-\tau)} \exp \bigg({-\frac{|\x - \x'|^2}{ 4 D_c (t-\tau)}} \bigg) \e^{\tau} \varrho_a(\x', \tau) \ \d \x' \ \d \tau.
$$
Thus the solution for the photormone dynamics after substituting $(t-\tau)=s$ and $t>0$, we have
\begin{equation}
c(\x, t) = \kh_\pm \e^{- t} \int_{0}^t \int_V \frac{1}{4 \pi \dh_c s} \exp \bigg({-\frac{|\x - \x'|^2}{ 4 \dh_c s}} \bigg) \e^{(t-s)} \varrho_a(\x', s) \ \d \x' \ \d s.
\end{equation}
If $\varrho_a = \alpha \delta (\x - \x')$, the above solution may be simplified to get
$$
c(\x, t) = \alpha \kh_\pm \int_{0}^t \frac{\e^{-s}}{4 \pi \dh_c s} \exp \bigg({-\frac{|\x|^2}{ 4 \dh_c s}} \bigg) \ \d s.
$$

\begin{table}[h!]
\centering
\begin{tabular}{ |c|l| }
\hline
\textbf{Length-scale} & \textbf{Process} \\
\hline
$l_a$ & Initial width of RAnt density \\
$D_a/v_o$ & RAnt density advection-diffusion \\
$(D_c/k_-)^{1/2}$ & Photormone diffusion-decay \\
\hline
\textbf{Time-scale} & \textbf{Process} \\
\hline
$\tau_a \sim l^2/D_a$ &  RAnt diffusion \\
$\tau_v \sim l/v_o$ & RAnt collective migration \\
$\tau_x \sim l^2/(\chi c_o)$ & Phototaxis \\
\hline
$\tau_+ \sim c_o/(k_+ \varrho_o)$ & Photormone production \\
$\tau_c \sim l^2/D_c$ & Photormone diffusion \\
$\tau_- \sim 1/k_-$ & Photormone decay \\
\hline
$\tau_s \sim 1/k_s$ & Substrate deposition \\
\hline
\end{tabular}
\caption{Length-scales and time-scales associated with different processes in the model in Eqs.~\ref{eq:SIrhor}-~\ref{eq:SIrhos}.} \label{tab:timeScale}
\end{table}

\section{Numerical simulation} \label{sec:simDet}
The 2D simulations of Eqs.~\ref{eq:SIndrhoa}-\ref{eq:SIndrhos} shown in Fig.~\TwoDsimul$(a)$ of the main text were performed in the general form Partial Differential Equations solver of commercial software $\textsf{COMSOL}^{\rm TM}$. The maximum mesh size was set to 0.1 in a domain of size $8 \times 6$ units$^2$. In the construction scenario, we chose an initial condition for the agent density, $\varrho_a(x,y,0)$ as $\Theta((x-x_o)^2 + (y-y_o)^2)^{1/2} - 1.0)\Theta((x-x_o)^2 + (y-y_o)^2)^{1/2} - 1.3)$ where $x_o = 5, y_o = 4$ and the density of the corral $\varrho_s(x,y,0)$ as $\Theta((x-x_o)^2 + (y-y_o)^2)^(1/2) - 0.5)$. In the de-construction scenario the initial densities were $\varrho_a(x,y,0) = \exp(-((x-x_o)^2+(y-y_o)^2)/2l_a^2)$ where $x_o=4, y_o=3.5, l_a=0.5$ and $\varrho_s(x,y,0) = \Theta(y-y_o)$ where $y_o=4.0$. We set the parameter $\varrho^* = 0.3, c^* = 0.01$ and the other parameters are set to $v_o = 0.1, \chi = 0.005, D_a = 0.005, k_+ = k_- = 1.5, D_p = 0.005$. We choose $k_s = 2.5$ for cooperative construction and $k_s = - 2.5$ for de-construction. A detailed analysis of the different possible phases of de-construction with the parameters in the model is analysed in~\cite{Prasath2021cooperative}.

\section{Robot design}\label{sec:robDes}

\begin{figure*}[h!]
\centering
\includegraphics[width=0.55\textwidth]{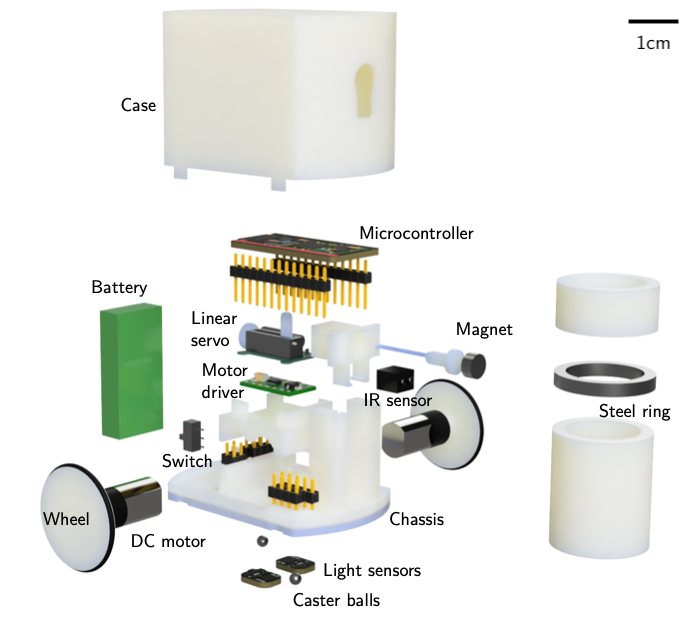}
\caption{Exploded view of a RAnt and a substrate element.}
\label{fig:ExpRAnt}
\end{figure*}

We designed the robots with the requirement that they are 1. mobile, 2. capable of sensing substrate elements, other RAnts, and the photormone field, and 3. capable of transporting substrate elements. An expanded view of the components inside each RAnt is shown in Fig. \ref{fig:ExpRAnt}. We used a rechargeable 3.7V battery with 400mAh (Pkcell LIPO 801735) and a Adafruit ItsyBitsy M0 Express microcontroller. The wheel diameter of the RAnts is 25mm and are driven by brushed DC motors rated at 85 RPM at 3.7V. Rubber o-rings increase the traction of the wheels. The motor speed is controlled with a Pololu DRV8835 Dual Motor Driver Carrier. Substrate elements are attached to and detached from the robots with a linear servomotor (Spektrum SPMSA2005) that extends and retracts a permanent magnet out of the robot's case. A ferromagnetic ring (3mm thickness colorFabb SteelFill) embedded in the substrate elements (22mm$\times$40mm PVC cylinders) allows for easy attachment to the permanent magnet when it is extended. RAnts are equipped with two light sensors (Adafruit ALS-PT19) located at the bottom left and right of the RAnt (relative to the direction of travel) and an infrared (IR) distance sensor (Everlight ITR20001 opto interrupter) for obstacle detection up to 3cm from the sensor. The base of the RAnt is 3D printed using acrylic styrene acrylonitrile (ASA). The wheel arrangement is inspired by the zooid robots \cite{le2016zooids} and requires two small steel caster balls of 3mm in diameter to be installed at the base that help stabilize the RAnt. The 3D printed case made of ASA protects the RAnt's internal components. A small switch is used to turn the power on/off with the case attached. Blue stickers of 6mm in diameter placed on top of the RAnts and red stickers on the substrate elements are used for position tracking with the webcam mounted above the arena.

\subsection{RAnt internal rules}
The internal rules of the RAnts are summarized in the pseudocode shown in Fig. 2 in the main text. The program uses a binary variable $d$ that encodes the direction of travel (1 for forward, -1 for backward), the cooperation parameter $\C\in[0,1]$, the deposition rate $\T\in[-1,1]$, the high threshold value $c_{h}=\bar{c}+\Delta c$, the low threshold value $c_{l}=\bar{c}-\Delta c$, and the maximal detectable photormone value $c_{max}$.

After the variables are defined, the program enters a while loop which is running until the RAnt is switched off or the battery voltage drops below 3.5V. The heading of the RAnt is set by adjusting the turning rate as a function of the cooperation parameter and the current sensor readings. Depending on the cooperation parameter, the turning rate is a composition of a random walk and phototaxis and follows the equation
\begin{equation}
\Omega = \frac{\C}{l_s} \tanh{\left(\alpha d \mathsf{K} \frac{c_L-c_R}{c_{max}} \right)} + \frac{\left( 1-\C \right)}{l_s} b \, \sin \left( \pi W \right)
\label{eq:turning}
\end{equation}

with $l_s=1cm$ the distance between the left and right sensors, $\alpha$ (=50) a gain parameter, $c_L$ and $c_R$ the photormone intensity measured in the left and right light sensors, respectively, $b=0.3$ a fixed amplitude, and a stochastic process $W$ (Wiener process). Using a sine function we map the stochastic process $W$ to the range $[-1,1]$ to avoid perpetual rotations for large excursions of $W$. The first term in Eq.~\ref{eq:turning} corresponds to phototaxis using the projected photormone and the second term to a random walk. We can tune the influence of either terms with the cooperation parameter $C$ from pure phototaxis at $C=1$ to a random walk at $C=0$. The deposition rate $\T$ defines the success of deposition and its sign determines wheter RAnts construct (positive sign) or de-construct (negative sign). The turning rate is used to determine the rotational speed of each wheel by
\begin{equation}
\omega_{L,R}=d\left(v_0 \mp G \frac{l_w}{2}\Omega\right),  \label{eq:wheelSpeed}
\end{equation}
with $v_0=4cm/s$ the base speed, $G=10^{-2}$ the rotational gain, and $l_w=3cm$ the distance between the two wheels. These parameters are chosen in a regime where self-trapping is not possible, but collective trapping is, therefore requiring multiple robots to be present to form nucleation sites.

After setting the new heading, the distance sensor is checked for any obstacles that are within $3 \cm$ in front of the RAnt. If an object is detected and the condition $\mathsf{K} \mathsf{C} \, c>\mathsf{K}(\bar{c}+\mathsf{K}\Delta c) $ is satisfied, the RAnt performs a fetching manoeuvre to attach to the substrate element. After the fetching manoeuvre, the direction parameter is inverted, i.e. $d=-1$. If an object is picked up with the magnet after the fetching manoeuvre, the distance sensor will report a detected object as long as it is attached to the magnet. Since $d=-1$, the RAnt will perform the same type of gradient driven locomotion described in Eq.~\ref{eq:turning} and Eq.~\ref{eq:wheelSpeed} but the sign of the signal sent to the motor driver will be inverted, resulting in a reverse motion of the RAnt. If an object is detected, but the photormone concentration does not satisfy the inequality, an avoidance manoeuvre is performed which consists of a random rotation in place in any direction with the intent to turn away from the detected obstacle.

The next if-statement checks if an obstacle is detected, but without the condition that the direction parameter is equal to one. If no obstacle is detected, the direction parameter $d$ is set to one and the magnet is disengaged. This increases the robustness of the system in case an obstacle is accidentally dropped.

The last if-statement checks whether the RAnt is in the reverse mode $d=-1$ and if the photormone concentration dropped satisfies the inequality $\mathsf{K} \, c < \mathsf{C} \mathsf{K}(\bar{c}-\mathsf{K}\Delta c) $. if both statements are true, the magnet is disengaged, depositing any potentially attached wall elements, and the direction parameter is set back to $d=1$. In order to avoid the RAnt from reattaching to the just dropped element, it performs a random rotation in place in any direction before returning to the start of the main loop.

\subsection{Experimental set-up}
The photormone was produced with an Epson EX9200 projector onto an acrylic sheet with a translucent top, which served as the surface on which the RAnts are operating. The projector uses three-chip digital light processing (DLP) which is required for the light sensors in the RAnts to pick up the photormone field. The dynamics of the photormone field is a function of the RAnt's positions and is given by
\begin{equation}
\partial_t c = - k_- c + k_+ \sum_{i=1}^n f(\bm{r}_{i},w)
\end{equation}
with $c=c(\bm{x},t)$ the photormone concentration at position $\bm{x}=[x,y]$ and time $t$, $k_-=0.02 \, \rm{s^{-1}}$ the decay rate, $k_+=0.1 \, \rm{s^{-1}}$ the photormone production rate, $n$ the number of RAnts detected in the arena, $f(\bm{r}_{i},w)$ a function with the property that it is zero everywhere except around a circle with diameter $w=2.5cm$ centered at the geometric center of the $i$th RAnt. The value of $f$ inside this circle is 1. The position of the RAnts are used as the centers of sources of photormone. If a RAnt is not moving, photormone is built up with rate $k_+$ at that location over time. When the RAnt moves to a new location, the built up photormone decays with rate $k_-$. The parameter choice described here has shown to neither saturate the domain with photormone nor be too volatile, but allowing the photormone to act as a spatiotemporal memory for the RAnts over the course of an experiment.

The positions of the RAnts are tracked with a webcam mounted above the arena and evaluated in \textsc{Matlab}. Blue markers are attached on the centroid of the case's upper surface which allow to use a simple blob detection to identify the pixel position of the RAnts. The photormone concentration is then dynamically updated in the same \textsc{Matlab} script and displayed on the RAnt arena with the projector. The tracking and integration of the photormone field is executed in real time which restricted the update rate of the projected field to $13 \, \rm{Hz}$ on average. The low refresh rate did not have any noticeable consequences for the conducted experiments.

\begin{figure}[ht!]
\centering
\includegraphics[width=0.8\textwidth]{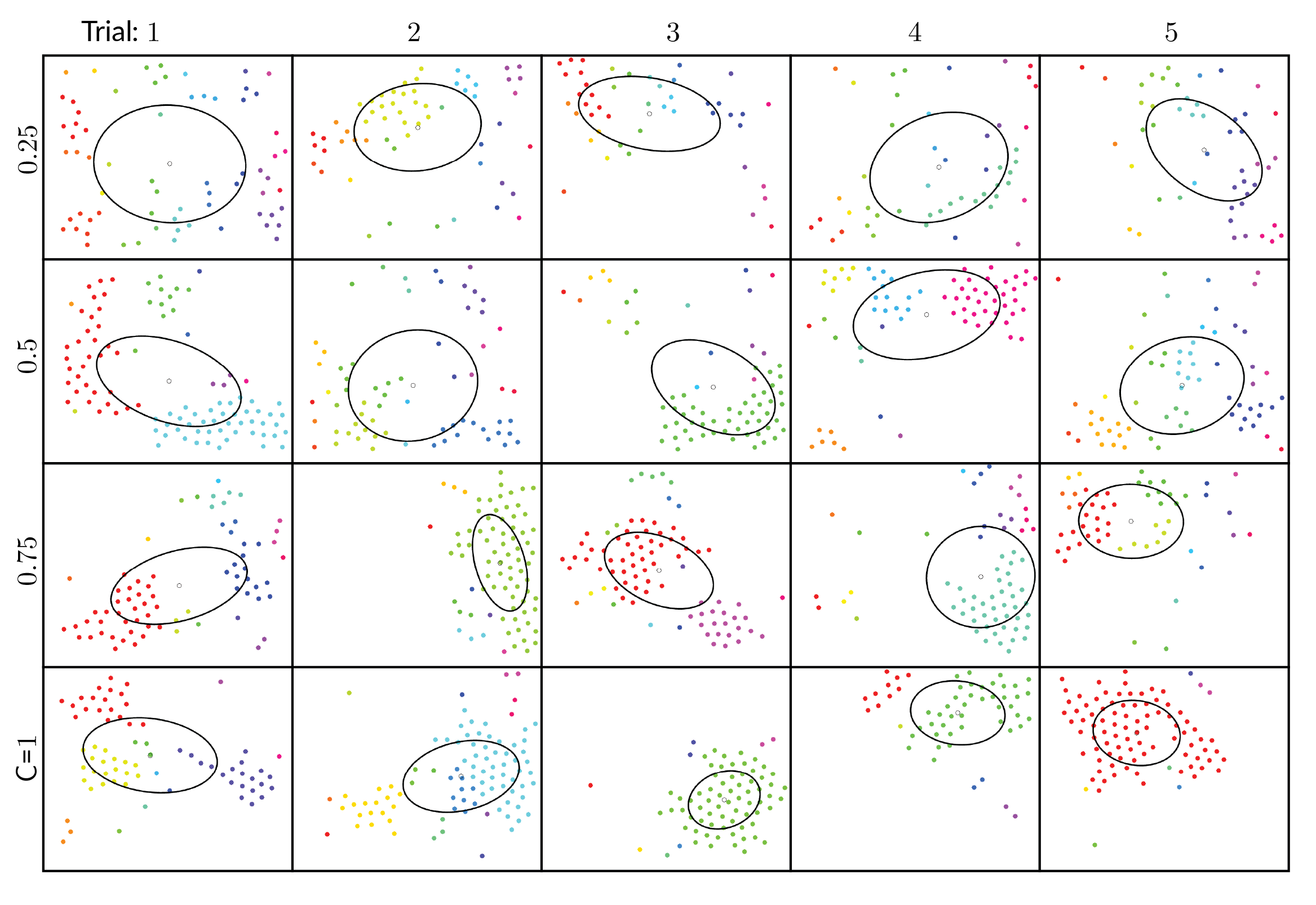}
\caption{Final substrate element position of all conducted robot construction experiments (see also Fig. \rantResults) for different values of cooperation parameter $\C$. For each cooperation parameter five trials were conducted with different initial positions of the robots. Same colored substrate elements belong to the same cohesive cluster such that element $i$'s centroid at least satisfies the condition $||\bm{x}_i-\bm{x}_j||<\delta$ ($\delta$=25mm) with another element $j$ in the cluster. The drawn ellipses represent the sample covariance of the substrate element distribution in the construction area, effectively capturing the cohesiveness of all elements in the arena. As the cooperation parameter increases, less clusters are observed (fewer colors), and the size of the obstacle distribution captured by the area of the ellipse also decreases.}
\label{fig:ExtendedResults}
\end{figure}

The set-up of the enclosure for the RAnts consisted of solid walls that contained the RAnts within an arena of $67 \times 56 cm^2$. $200$ substrate elements were distributed along the boundary of the arena and served as the construction material. We tracked all substrate elements by attaching a red dot on each of them and tracking them with the webcam. For every experiment we randomly placed the rants in the arena and waited for the limit of 10 minutes to be reached. In some cases we stopped the experiment early due to fast excavation and depletion of the substrate element layer. At that point, data was stored and the experiment ended. There was no leader and no dedicated roles, which makes every RAnt replaceable. The construction area, that is, the area in which we projected photormone, was $48 \times 35 cm^2$. The smaller size ensured that a distinction between the initial substrate element locations and constructed structures can be made. Every element inside the construction area will have had to be moved there by a RAnt.

We conducted experiments for five cooperation parameters $\C=\{0,0.25,0.5,0.75,1\}$ at fixed excavation rate $\T=1$ and repeated experiments five times for each parameter. Every RAnt's software was updated before a new set of five experiments with the same cooperation parameter was conducted. The projected photormone field was contained in an area smaller than the full arena to ensure that the RAnts start constructing away from the boundary. For every experiment, we stored the webcam data and time stamps. The video frames were post-processed and locations of all RAnts and wall elements were stored as a function of time.

For the de-construction experiment we used $\C=1$ and $\T=-1$. The only thing that need to be changed in the RAnts code were the values of these two parameters. We arranged the substrate elements along the longer edge of the arena, forming 7 layers through which the RAnts can excavate before reaching the boundary. Since de-construction requires an existing substrate element to be excavated at locations of high photormone concentration (which may or may not happen naturally), we initialized the photormone field with a small ($~4cm$ diameter) seed at the location of the boundary that decays away at the same rate as regular photormone. This was enough to encourage excavation at the seed's location. We ran the excavation experiment until the first RAnt reached the solid boundary of the arena.

\section{Extended experimental results}
The final substrate element positions of the main experiments as a function of the cooperation parameter are shown in \ref{fig:ExtendedResults}. Substrate elements of same color belong to the same cluster. Two substrate elements belong to the same cluster if the distance of their centroids satisfies $||\bm{x}_i-\bm{x}_j||<\delta$ ($\delta$=25mm). The ellipse indicates the sample covariance of the substrate elements located in the construction area.

\begin{figure}[ht!]
\centering
\includegraphics[width=0.7\textwidth]{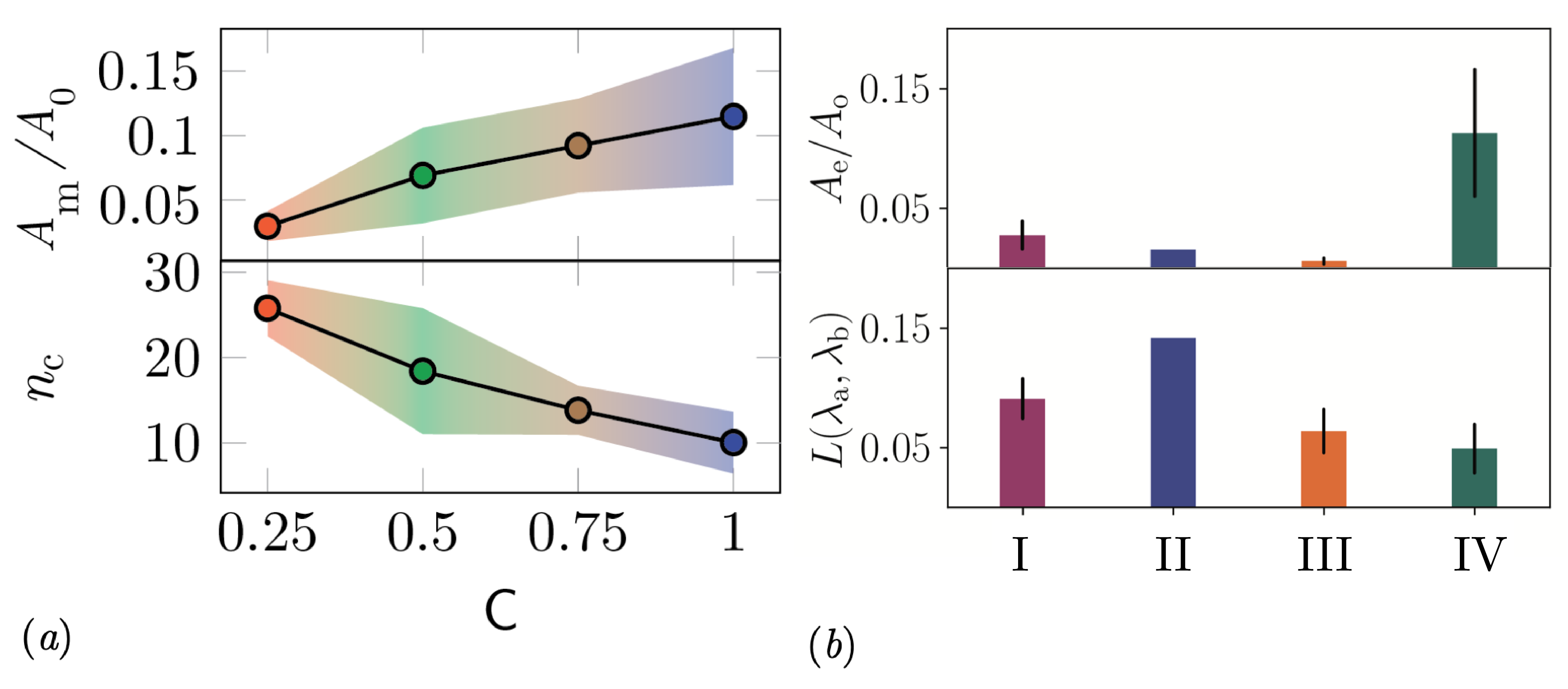}
\caption{Extended experimental results of robot construction. $\bm{(a)}$ Robotic construction at different cooperation parameters $\C$ (five trials per $\C$), extending results shown in Fig.~\rantResults and \ref{fig:ExtendedResults}. \textbf{Top}: Covered relative area $A_m/A_0$ of substrate in the construction arena. $A_m$ is the area covered by the largest cluster and $A_0$ is the size of the construction area. The higher the cooperation parameter, the larger the fraction of the total area covered by the biggest cluster. \textbf{Bottom}: number of clusters in construction area, $n_c$. With increasing cooperation parameter the number of clusters decreases.  $\bm{(b)}$ Extended results of construction robustness experiments shown in Fig.~\TwoDsimul$(b)$. Four cases of collective construction are studied: (I) with no threshold and phototaxis, $c^* = 0, \C=0$;  (II) with no threshold and strong phototaxis,  $c^* = 0, \C=1$; (III) with threshold and strong phototaxis, $c^* \neq 0, \C=1$; (IV) with threshold and no phototaxis, $c^* \neq 0, \C=0$. \textbf{Top}: total covered area of deposited substrate $A_\text{e}$ (normalized by construction arena area $A_\o$) of constructed structure. \textbf{Bottom}: circumference $L$ of covariance ellipses as in Fig.~\TwoDsimul$(b)$ of deposited substrate elements.}
\label{fig:clusterNumber}
\end{figure}
An extension of the results shown in Fig. \rantResults is shown in Fig. \ref{fig:clusterNumber}$(a)$, where the relative area of the largest cluster and the number of clusters is shown as a function of the cooperation parameter $\C$. The area of the largest cluster, $A_m$, increases with the cooperation parameter, indicating an increased cohesiveness of the formed structures. The number of clusters, $n_c$, on the other hand decreases with cooperation parameter, which indicates fewer stray substrate elements due to increased cohesiveness.
\end{document}